\documentclass[10pt,twocolumn,letterpaper]{article}

\usepackage{cvpr}              

\usepackage[table,dvipsnames]{xcolor} 

\usepackage{comment}
\usepackage{amsfonts}
\usepackage{amsmath}
\usepackage{ragged2e}
\usepackage{booktabs}
\usepackage{multirow}
\usepackage{graphicx}

\usepackage{colortbl}
\usepackage{array}
\usepackage{adjustbox}
\usepackage{enumitem}

\definecolor{grey}{rgb}{0.93,0.93,0.93}
\definecolor{lightlightgrey}{rgb}{0.9,0.9,0.9}
\definecolor{lightlighblue}{rgb}{0.8,0.89,0.93}
\definecolor{pastel_blue}{rgb}{0.7,0.85,0.95}
\definecolor{pastel_green}{rgb}{0.7,0.9,0.7}
\definecolor{pastel_orange}{rgb}{0.98,0.78,0.59}

\usepackage{tcolorbox}
\newtcolorbox{mybox}[2]{
    arc=3pt,
    boxrule=#2pt,
    colback=#1,
    width=8.25cm,
    halign=left,
}

\definecolor{cvprblue}{rgb}{0.21,0.49,0.74}
\usepackage[pagebackref,breaklinks,colorlinks,allcolors=cvprblue]{hyperref}

\def\confName{CVPR}
\def\confYear{2026}

\title{
Thinking Beyond Labels: \\
Vocabulary‑Free Fine‑Grained Recognition using Reasoning-Augmented LMMs
}

\author{
\normalsize Dmitry Demidov \\
{\tt\small dmitry.demidov@mbzuai.ac.ae}
\and
\normalsize Zaigham Zaheer \\
{\tt\small zaigham.zaheer@mbzuai.ac.ae}
\and
\normalsize Zongyan Han \\
{\tt\small zongyan.han@mbzuai.ac.ae}
\and
\normalsize Omkar Thawakar \\
{\tt\small omkar.thawakar@mbzuai.ac.ae}
\and
\normalsize Rao Anwer \\
{\tt\small rao.anwer@mbzuai.ac.ae}
\\
\vspace{-15.0pt}
\and
\small Mohamed bin Zayed University of Artificial Intelligence, UAE 
}

\begin{document}

\twocolumn[{%
\renewcommand\twocolumn[1][]{#1}%
\maketitle
\vspace{-1.2em}
\centering
\includegraphics[width=0.95\textwidth]{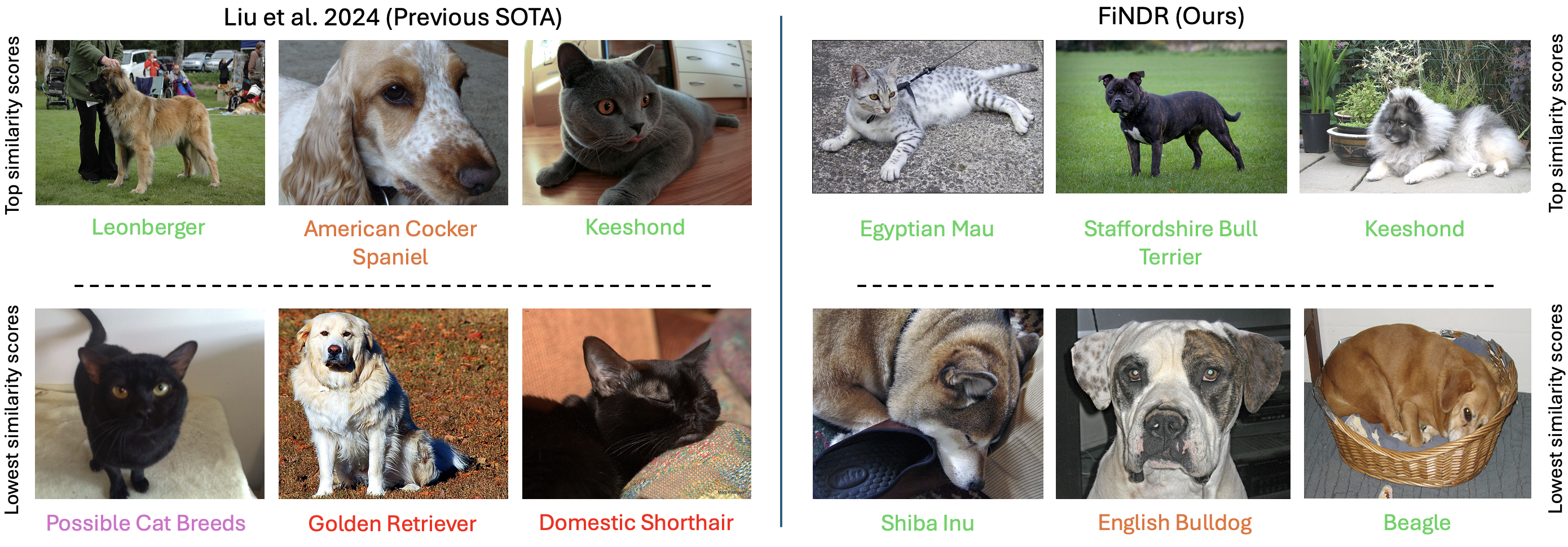} 
\vspace{-0.5em}
\captionof{figure}{ 
Qualitative comparison of a predicted label vocabulary between the previous state-of-the-art method \cite{finer} 
and our approach (FiNDR) on the Oxford Pets dataset.
The predicted class names are obtained for the unlabelled discovery set. All predictions are sorted by the similarity score with top-3 and bottom-3 labels depicted for each method.
\textit{Legend}: predictions in \textcolor{LimeGreen}{green} are correct, predictions in \textcolor{orange}{orange} are partially correct, predictions in \textcolor{red}{red} are incorrect, predictions in \textcolor{magenta}{pink} are failed.
\vspace{1.5em}
}
\label{fig:analysis_vocabulary}
}]
\begin{abstract}
%
%
%
Vocabulary‑free fine-grained image recognition aims to distinguish visually similar categories within a meta-class without a fixed, human‑defined label set.
%
Existing solutions for this problem are limited by either the usage of a large and rigid list of vocabularies or by the dependency on complex pipelines with fragile heuristics where errors propagate across stages.
%
%
%
Meanwhile, the ability of recent large multi‑modal models (LMMs) equipped with explicit or implicit reasoning to comprehend visual-language data, decompose problems, retrieve latent knowledge, and self‑correct suggests a more principled and effective alternative.
%
%
%
Building on these capabilities, we propose FiNDR (\underline{Fi}ne‑grained \underline{N}ame \underline{D}iscovery via \underline{R}easoning), the first reasoning‑augmented LMM-based framework for vocabulary‑free fine‑grained recognition. 
The system operates in three automated steps: (i) a reasoning‑enabled LMM generates descriptive candidate labels for each image; (ii) a vision-language model filters and ranks these candidates to form a coherent class set; and (iii) the verified names instantiate a lightweight multi‑modal classifier used at inference time. 
%
%
Extensive experiments on popular fine-grained classification benchmarks demonstrate state-of-the-art performance under the vocabulary-free setting, with a significant relative margin of up to 18.8\% over previous approaches. 
Remarkably, the proposed method surpasses zero‑shot baselines that exploit pre-defined ground‑truth names, challenging the assumption that human‑curated vocabularies define an upper bound.  
Ablations further confirm that advanced 
prompting techniques and built-in reasoning mechanisms significantly enhance naming quality. 
Additionally, we show that carefully curated prompts enable open‑source LMMs to match proprietary counterparts.
These findings establish reasoning‑augmented LMMs as an effective foundation for scalable, fully automated, open‑world fine‑grained visual recognition.
The source code 
is available on \href{https://github.com/demidovd98/FiNDR}{github.com/demidovd98/FiNDR}.
%
%
\end{abstract}    
\section{Introduction}
\label{sec:intro}


Fine-grained image classification aims to distinguish visually similar categories by leveraging subtle yet discriminative details. While conventional fine-grained recognition methods \cite{Wang2021FeatureFV,lagunas2023transfer, chou2022novel} have achieved significant progress, they often depend heavily on 
extensive vocabularies pre-defined by domain experts. This reliance inherently limits their generalization applicability in open-world scenarios, where prior knowledge about the domain-specific categories might be incomplete or entirely unavailable. 
Consequently, the vocabulary-free fine-grained recognition task, which involves classifying images without any predetermined categorical vocabulary, has emerged as an essential yet challenging direction \cite{conti2024vocabularyfreeimageclassification,finer}.
Current vocabulary-free methods generally fall into three major categories: clustering-based methods, zero-shot methods with predefined vocabularies, and dynamic vocabulary discovery using vision-language models (VLMs) or large language models (LLMs). Clustering-based methods~\cite{krishna1999genetic,chen2022plot} rely solely on visual features, which limits their practicality, particularly when object name identification is needed.
While zero-shot learning~\cite{parikh2011relative,akata2015evaluation,han2020learning,han2021contrastive} exhibits promising generalisation, it remains dependent on large, predefined vocabularies. This reliance becomes a major limitation when label information is missing, noisy, or misaligned. 
To solve this, methods employing dynamic vocabulary discovery \cite{finer,e-finer} utilise a small set of unlabelled images to discover the potential class names, assuming the complex low-data scenario \cite{schmarje2021survey, xu2021end}. 
However, it further decouples the recognition process into multiple independent stages involving different models, introducing potential gaps and inconsistencies. Moreover, the attributes generated by LLMs are often not image-specific, making them less reliable, especially in the presence of image variance and intra-class diversity.



%
Meanwhile, the most recent advancements in foundational Large Multi-modal Models (LMMs), trained on massive 
datasets, offer significant potential to address these limitations. 
These LMMs, traditionally combining vision and language modalities, have demonstrated exceptional generalisation capabilities across diverse tasks without explicit domain-specific training, often outperforming specialised ``one-task, one-model" approaches by utilising their broad knowledge and flexible capabilities \cite{li2023blip2,dai2023instructblip,liu2023llava,qwen2vl2024,bai2024internvl2}.
%
%
%
Another crucial advancement is the emergence of LMMs equipped with reasoning mechanisms, allowing them to mitigate complex tasks by facilitating deeper semantic comprehension and enabling dynamic inference.
Specifically, recent LMMs incorporating explicit or implicit chain-of-thought style processing have shown the capabilities to decompose problems, dynamically retrieve latent knowledge based on context, and even self-correct their outputs. \citep{wei2022cot,kojima2022zero_shot_reasoners,wang2023self_consistency}.
%
%
%
Such integrated reasoning is particularly well-suited for vocabulary-free fine-grained recognition, which demands nuanced visual discrimination and contextual knowledge without dependency on extensive predefined vocabularies.
While these reasoning-specialised LMMs have previously been dominated by the private solutions, multiple public counterparts with on-par performance have recently become available \cite{bai2025qwen25vltechnicalreport, zhu2025internvl3exploringadvancedtraining}.
However, until now, the application of these advanced LMMs to the vocabulary-free fine-grained setting has remained largely unexplored.


%
In this work, we investigate whether the vast knowledge embedded in modern LMMs, augmented with reasoning techniques, can be harnessed to construct an effective, fully automated system for vocabulary-free fine-grained recognition.
We propose a simple yet powerful novel pipeline that requires no prior category knowledge. Specifically, given a set of unlabelled images, a reasoning-enabled LMM is prompted to generate a set of descriptive candidate class names for each sample. Next, a highly-capable vision-language model filters and ranks these candidates to produce the final list of categories. Lastly, these class names are further used to build an advanced multi-modal classifier based on a computationally-friendly VLM utilised at inference time. 
Such a fully-automated approach capitalises on the rich implicit knowledge of foundation models while using the VLM as a semantic verifier, thereby avoiding any fixed vocabulary or human-curated priors, facilitating its use in genuinely open-world scenarios.

In extensive experiments across multiple fine-grained benchmarks,
our method achieves state-of-the-art accuracy in the vocabulary-free setting, substantially outperforming previous approaches.
Remarkably, it even surpasses a zero-shot classifier that uses the ground-truth class names as its vocabulary, which has traditionally been regarded as an unbeatable ``upper-bound" baseline for the vocabulary-free task. 
Our findings 
challenge the long-held assumption that human-designed, rigid vocabularies represent the most optimal choice for fine-grained recognition tasks, 
opening avenues for more adaptive, efficient, and automated visual classification systems.
We further conduct detailed analysis and ablation studies to elicit high-quality responses from LMMs. 
Additionally, we demonstrate that with carefully crafted prompts, open-source LMMs can match the performance of their proprietary, closed-source counterparts on this task, making high-quality vocabulary-free recognition accessible without private and pay-walled services.

Our main contributions are summarised as follows:
\begin{itemize}[
topsep=0pt, 
leftmargin=10pt
]
    \item To our knowledge, this is 
    the first study investigating the application of reasoning-augmented large multi-modal models for vocabulary-free fine-grained visual recognition, filling up an important gap in the literature. 
    
    \item We introduce a fully-automated novel framework that achieves 
    state-of-the-art results with a significant relative margin of up to 18.8\% over the previous methods,
    without any prior knowledge or pre-defined fixed vocabulary. 
    
    \item Remarkably, 
    our FiNDR 
    also outperforms zero-shot classifiers, which were previously considered unreachable ``upper-bound" baselines due to their access to known ground-truth human-specified class names.
    This important discovery challenges the presumed superiority of human-oriented labels in fine-grained recognition. 

    \item We provide an
    in-depth analysis of prompt design techniques and reasoning strategies across state-of-the-art public and private LMMs, offering practical, actionable guidelines that enable open-source models to achieve competitive performance with proprietary counterparts. 

\end{itemize}

\section{Related Work}
\label{sec:related-work}

\subsection{Vocabulary‑free fine‑grained recognition}

Vocabulary‑free fine‑grained recognition remains relatively under‑explored, with prior studies falling into three principal streams: 
(i) clustering‑based techniques, 
(ii) zero‑shot approaches that rely on predefined vocabularies, 
and (iii) methods that dynamically derive vocabularies through the integration of vision‑language models and large language models.
Clustering-based approaches operate directly on visual embeddings extracted from large pre-trained encoders (e.g., CLIP \cite{radford2021clip}, DINO \cite{caron2021dino}) and then partition features using algorithms such as KMeans \citep{krishna1999genetic} or Sinkhorn-Knopp variants \citep{chen2022plot,vaze2022gcd}. While conceptually simple, these methods return index-based clusters with no semantic grounding and exhibit unstable performance when tasked with subtle category boundaries. 
Zero-shot classification with a fixed vocabulary relies on a large but rigid label vocabulary, for instance, CLIP prompted with WordNet synsets or domain-specific taxonomies, as in SCD or CaSED \citep{minderer2022simple,zhang2023cased}. Such strategies broaden coverage beyond a small supervised set, however, they still require prior knowledge of possible classes and consequently struggle in genuinely open-world deployments. 
More recently, researchers have explored dynamic discovery of the vocabulary for each image using powerful language or vision-language models. For instance, FineR \cite{finer} generates class names on the fly by first describing the image with a vision model and then reasoning with a large language model to produce a fine-grained label. This removes the need for any predetermined list of labels but introduces its own challenges: such multi-stage pipelines are complex and prone to error propagation, as mistakes from one component affect the next. 
A derivative work, E-FineR \cite{e-finer}, proposes utilising an LLM for class-specific contextual grounding for the discovered labels, which further improves overall performance.
Nevertheless, such methods with complicated pipelines tend to generate confident inaccuracies while also relying on text-only models with limited training knowledge bases, which can lead to sub-optimal naming (e.g., heuristic or biased name choices). In fact, to date, these dynamic methods have not surpassed the accuracy of simpler zero-shot classifiers with known pre-defined class names, which are considered an unreachable upper-bound, highlighting the difficulty of the task.
Additional details and analysis of more related work can be found in App. \ref{app:extra_related_work}.

\section{Method}
\label{sec:method}

\begin{figure*}[!ht]
    \centering
    \includegraphics[width=1.0\textwidth]{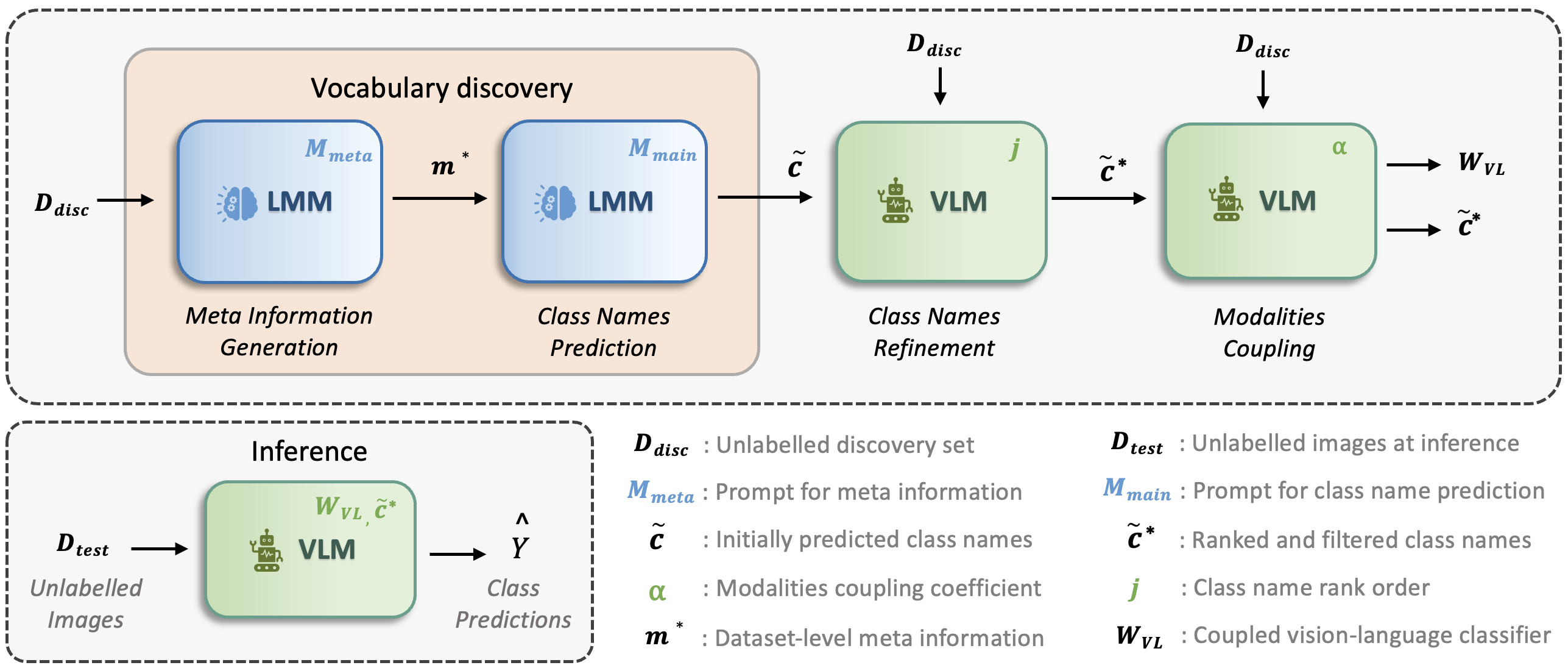}
    \caption{Overview of our FiNDR framework for vocabulary-free fine-grained image classification. The pipeline operates in two primary phases: \textit{vocabulary discovery} and \textit{classifier preparation}. In the \colorbox{pastel_orange}{vocabulary discovery} phase, a 
    large multi-modal model~(\colorbox{pastel_blue}{LMM}) generates dataset-level meta information from an unlabelled discovery set ($D_{\text{disc}}$), followed by
    initial class name predictions ($\tilde{c}$). 
    During classifier preparation, the class names are refined with a vision-language model~(\colorbox{pastel_green}{VLM}), which ranks and filters these candidates to produce a refined vocabulary set ($\tilde{c}^{*}$). Then, a modalities coupling step combines visual and textual embeddings to create a unified vision-language classifier ($W_{\text{VL}}$). 
    At inference time, this classifier is used to assign interpretable and fine-grained semantic labels to unseen test images ($D_{\text{test}}$), without relying on any predefined vocabulary.
    }
    \label{fig:main_architecture}
    \vspace{-1.0em}
\end{figure*}

\subsection{Problem Formulation}

In this work, we address fine-grained visual recognition in a vocabulary‑free setting.
To discover the potential class names, we assume an initial discovery set comprising a very small number of \(N\) unlabelled images: 
\(
\mathcal{D}_{\mathrm{disc}}
    =\{x_{i}\}_{i=1}^{N}\subset\mathbb{R}^{H\times W\times C}
\). 
It may be noted that no predefined dataset class tags, attribute metadata, or auxiliary supervision is utilized at any step of our framework. 
%

During evaluation, a query image \(x \in \mathcal{D}_{\mathrm{test}}\) must be assigned a semantic label \( c \in \mathcal{C} \), where the vocabulary \(\mathcal{C}\), a set of ground-truth fine‑grained names, is unknown a~priori.
Consequently, a viable solution must jointly 
(i) induce a plausible label list \(\mathcal{C}\) using only the visual information available in \(\mathcal{D}_{\mathrm{disc}}\), 
and (ii) define a mapping \(f : \mathcal{D}_{\mathrm{test}} \!\to\! \mathcal{C}\) that selects the most appropriate name for each query instance.
This dual objective 
fundamentally differentiates the vocabulary‑free paradigm from conventional zero‑shot learning, where the target label set is provided in advance.

\subsection{Fine‑grained Name Discovery via Reasoning}
In this subsection, we describe the details of our proposed Fine‑grained Name Discovery via Reasoning (FiNDR), with its overall architecture depicted in Figure \ref{fig:main_architecture}.

\subsubsection{Vocabulary Discovery via Reasoning.} 
\label{sec:vocab_discovery}

First, in order to achieve higher-quality class name predictions, we leverage the common technique of step-by-step prompting, beneficial for reasoning-specialised LMMs \cite{wei2022cot,bai2025qwen25vltechnicalreport,zhou2025visualizingstepreasoningmira}.
Specifically, for every image \(x_i\) in the discovery split \(\mathcal{D}_{\mathrm{disc}}\), the large multi-modal model is requested in two consecutive queries to first predict meta-information about the object in the image and then a candidate class name for this object.
Let $\mathcal{M}_{\mathrm{meta}}$ and $\mathcal{M}_{\mathrm{main}}$ be instantiations of the LMM $\mathcal{M}$ that differ only in their prompts (for details, refer to Appendix \ref{app:prompts_details}). 
We first obtain dataset‑level meta‑information by querying the model with a small, randomly sampled context set \(\mathcal{S}=\{x_{a},x_{b},x_{c}\}\subset D_{\text{disc}}\) of three images:
%
\begin{equation}
  m^{\star}
  \;=\;
  \mathcal{M}_{\mathrm{meta}}(\mathcal{S})
  \;=\;
  \bigl(c^{\text{meta}},\; u^{\text{type}},\; e^{\text{expert}}\bigr),
\end{equation}
where the prompt carried by the variant \(\mathcal{M}_{\mathrm{meta}}\) requests  
\textit{(i)} a broad taxonomic group \(c^{\text{meta}}\),  
\textit{(ii)} the granularity unit within that group \(u^{\text{type}}\), and  
\textit{(iii)} a domain specialist name \(e^{\text{expert}}\) needed for fine discrimination.
Given this frozen context, each individual image \(x_i\) is then passed, together with \(m^{\star}\), to the same backbone model under a second prompt that requests for the unique fine‑grained label:
%
\begin{equation}
  \tilde{c}_i
  \;=\;
  \mathcal{M}_{\mathrm{main}}\!\bigl(x_i,\, m^{\star}\bigr),
\end{equation}
where $\tilde{c}_i$ belongs to an initial set of candidate class names \( \tilde{\mathcal{C}} \), and \(\mathcal{M}_{\mathrm{main}}\) differs from \(\mathcal{M}_{\mathrm{meta}}\) only in its instruction text.
For specific prompts, refer to Appendix \ref{app:prompts_details}.

Finally, 
every raw prediction \(\tilde{c}_i\) is standardised and to discard unusable outputs.  
This post‑processing normalises whitespace, capitalisation, and pluralisation. Additionally, it filters out syntactically corrupted strings and rejects names that remain overly generic after normalisation.  
The final, cleaned labels constitute the induced vocabulary $\tilde{\mathcal{C}}$.

\subsubsection{Class Names Refinement.}
\label{sec:class_refinement}

While the LLM supplies a set of seemingly valid labels stored in \( \tilde{\mathcal{C}} \), some of these names are not truly representative of the images in our collection. To this end, we further employ a vision–language encoder such as CLIP to align text and image domains by comparing the text embeddings of \( \tilde{\mathcal{C}} \) with the visual features extracted from \( \mathcal{D}_{\text{disc}} \).

For every candidate label \(c \in \tilde{\mathcal{C}}\) we compute a visual relevance score via its average cosine similarity to each image embedding \(\mathbf{v}_j \in D_{\text{disc}}\):
%
\begin{equation}
    \text{score}(c) = \frac{1}{N} \sum_{j=1}^{N} \cos(\mathbf{t}_c, \mathbf{v}_j) = \frac{1}{N} \sum_{j=1}^{N} \frac{\mathbf{t}_c^\top \mathbf{v}_j}{\|\mathbf{t}_c\| \cdot \|\mathbf{v}_j\|} .
\end{equation}

Ranking the candidates by this score and retaining only the highest‑scoring entries yields the refined vocabulary \(\tilde{\mathcal{C}^*} \subseteq \tilde{\mathcal{C}}\).  This filtering step narrows the semantic space of textual labels to those that best match the visual evidence present in the dataset.

\subsubsection{Vision-Language Modalities Coupling.}
\label{sec:vl_coupling}

To further boost similarity scores among images belonging to the same category, we couple their textual and visual embeddings \cite{finer}.  Crucially, this stage relies solely on automatically assigned labels, requiring zero manual involvement.

Using the refined vocabulary \( \tilde{C^*} \), we first create a text classifier for every label \( c \in \tilde{C^*} \) by encoding the name with the text branch of our vision‑language model. 
Since a name alone can still be ambiguous \cite{radford2021clip}, we complement these text prototypes with visual information.  Each image in the discovery split \( D_{disc} \) is pseudo‑labelled by choosing the class in \( \tilde{C^*} \) whose text embedding yields the highest cosine similarity, producing a set \(\mathcal{U}_c\) of \( U_c \) pseudo‑labelled samples per class.  Since \( D_{disc} \) provides only a handful of examples for each label, the resulting visual features are prone to bias and a lack of diversity. To alleviate this, we apply a simple augmentation pipeline with random crops and horizontal flips applied \( K = 10 \) times to every image. 
The augmented features are then averaged to build a visual classifier \(\mathbf{v}_c\):
%
\begin{equation}
    \mathbf{v}_c = \frac{1}{K \cdot |\mathcal{U}_c|} \sum_{i=1}^{|\mathcal{U}_c|} \sum_{k=1}^{K} \frac{f_V(\text{Aug}_k(\mathbf{x}_i^c))}{\|f_V(\text{Aug}_k(\mathbf{x}_i^c))\|},
\end{equation}
\noindent where \(\text{Aug}_k\) denotes the \(k\)-th augmentation and \(f_V\) is the visual encoder of the vision‑language model.  Restricting ourselves to the same two augmentations used when training CLIP models \cite{CLIP} keeps the resulting embeddings well aligned with those encountered at test time.

Finally, text and visual representations are merged into a single vision–language classifier for each class:
%
\begin{equation}
    W_{VL}^{(c)} = \alpha \cdot \mathbf{t}_c + (1 - \alpha) \cdot \mathbf{v}_c ,
\end{equation}
\noindent where \(\alpha\) is a coupling coefficient fixed to \( \alpha = 0.7 \) throughout our experiments (for ablation, refer to Appendix \ref{app:alpha_ablation}). This vision-language mixing allows the model to mitigate the potential domain shift and noisiness of guessed textual class names by taking advantage of the complementary strengths of both modalities (for details, see App. \ref{app:domain_shift}).

\subsubsection{Inference.}
\label{sec:inference}

At inference, given a previously unseen image \( x \in \mathcal{D}_{\text{test}} \), the model extracts its visual embedding, which is used to assign a class by comparing this vector against every coupled prototype in \( \mathcal{W}_{\text{VL}} \) drawn from the refined label set \( \tilde{\mathcal{C}}^* \):
%
\begin{equation}
    \tilde{y} = \arg\max_{c \in \tilde{\mathcal{C}^*}} \cos\bigl( f_V(\mathbf{x}),\, W_{VL}^{(c)} \bigr) ,
\end{equation}
\noindent where the cosine metric quantifies the alignment between the test image embedding and each vision–language class representation.
Importantly, the predicted output \( \tilde{y} \) is a human‑readable semantic name rather than a numeric index, which enhances interpretability and facilitates deployment in practical settings. 

In this way, the entire proposed pipeline, from discovering class names to the final prediction, operates without any manual supervision, training, or pre-defined vocabularies.
\section{Experiments and Analysis}

\begin{table*}[!h]
    \centering
    \normalsize
    \setlength{\tabcolsep}{4pt}
    \renewcommand{\arraystretch}{1.1}
    \begin{adjustbox}{max width=\textwidth}
    \begin{tabular}{l|cc|cc|cc|cc|cc|cc}
    \toprule
    \textbf{} & \multicolumn{2}{c|}{\textbf{Birds-200}} & \multicolumn{2}{c|}{\textbf{Cars-196}} & \multicolumn{2}{c|}{\textbf{Dogs-120}} & \multicolumn{2}{c|}{\textbf{Flowers-102}} & \multicolumn{2}{c|}{\textbf{Pets-37}} & \multicolumn{2}{c}{\textbf{Average}} \\
    \textbf{Method} & cACC & sACC & cACC & sACC & cACC & sACC & cACC & sACC & cACC & sACC & cACC & sACC \\
    \midrule
    \textcolor{gray!70}{CLIP (zero-shot) \cite{CLIP}} & \textcolor{gray!70}{57.4} & \textcolor{gray!70}{80.5} & \textcolor{gray!70}{63.1} & \textcolor{gray!70}{66.3} & \textcolor{gray!70}{56.9} & \textcolor{gray!70}{75.5} & \textcolor{gray!70}{69.7} & \textcolor{gray!70}{77.8} & \textcolor{gray!70}{81.7} & \textcolor{gray!70}{87.8} & \textcolor{gray!70}{65.8} & \textcolor{gray!70}{77.6} \\
    \midrule
    \multicolumn{13}{c}{\textit{Index-based approaches without vocabulary}} \\
    \midrule
    CLIP-Sinkhorn \cite{chen2022plot} & 23.5 & - & 18.1 & - & 12.6 & - & 30.9 & - & 23.1 & - & 21.6 & - \\
    DINO-Sinkhorn \cite{chen2022plot} & 13.5 & - & 7.4 & - & 11.2 & - & 17.9 & - & 5.2 & - & 19.1 & - \\
    KMeans \cite{krishna1999genetic} & 36.6 & - & 30.6 & - & 16.4 & - & 66.9 & - & 32.8 & - & 36.7 & - \\
    \midrule
    \multicolumn{13}{c}{\textit{Approaches with pre-defined vocabulary}} \\
    \midrule    
    WordNet \cite{WordNet} & 39.3 & 57.7 & 18.3 & 33.3 & 53.9 & \underline{70.6} & 42.1 & 49.8 & 55.4 & 61.9 & 41.8 & 54.7 \\
    CLEVER \cite{DBLP:journals/corr/abs-2111-03651} & 7.9 & - & - & - & - & - & 6.2 & - & - & - & - & - \\
    SCD \cite{han2024whatsnameclassindices} & 46.5 & - & - & - & \underline{57.9} & - & - & - & - & - & - & - \\
    CaSED \cite{conti2024vocabularyfreeimageclassification} & 25.6 & 50.1 & 26.9 & 41.4 & 38.0 & 55.9 & \underline{67.2} & 52.3 & 60.9 & 63.6 & 43.7 & 52.6 \\
    \midrule
    \multicolumn{13}{c}{\textit{Approaches with automatically discovered vocabulary}} \\
    \midrule    
    BLIP-2 \cite{blip-2} & 30.9 & 56.8 & 43.1 & 57.9 & 39.0 & 58.6 & 61.9 & \textbf{59.1} & 61.3 & 60.5 & 47.2 & 58.6 \\    
    FineR \cite{finer} & 51.1 & 69.5 & 49.2 & 63.5 & 48.1 & 64.9 & 63.8 & 51.3 & \underline{72.9} & 72.4 & 57.0 & 64.3 \\
    E-FineR \cite{e-finer} & \underline{52.1} & \underline{70.1} & \underline{51.2} & \underline{64.0} & 51.8 & 67.1 & 64.8 & 54.0 & 71.7 & \underline{76.2} & \underline{58.4} & \underline{66.3} \\
    \midrule
    \rowcolor{cyan!10}
    \textit{FiNDR (Ours)} & \textbf{58.8} & \textbf{76.9} & \textbf{56.1} & \textbf{64.4} & \textbf{58.1} & \textbf{71.0} & \textbf{79.8} & \underline{56.5} & \textbf{86.5} & \textbf{83.7} & \textbf{67.9} & \textbf{70.6} \\
    $\Delta$ to previous SOTA, absolute \% & \cellcolor{green!15}+6.7 \% & \cellcolor{green!15}+6.8 \% & 
    \cellcolor{green!15}+4.9 \% & \cellcolor{green!15}+0.4 \% & \cellcolor{green!15}+0.2 \% & \cellcolor{green!15}+0.4 \% & \cellcolor{green!15}+12.6 \% & \cellcolor{green!05}-2.6 \% & 
    \cellcolor{green!15}+13.6 \% &
    \cellcolor{green!15}+7.5 \% & \cellcolor{green!15}+9.5 \% & \cellcolor{green!15}+4.3 \% \\
    $\Delta$ to previous SOTA, relative \% & \cellcolor{green!15}+12.9 \% & \cellcolor{green!15}+9.7 \% & 
    \cellcolor{green!15}+9.6 \% & \cellcolor{green!15}+0.7 \% & \cellcolor{green!15}+0.4 \% & \cellcolor{green!15}+0.6 \% & \cellcolor{green!15}+18.8 \% & \cellcolor{green!05}-4.3 \% & 
    \cellcolor{green!15}+18.7 \% &
    \cellcolor{green!15}+9.9 \% & \cellcolor{green!15}+16.3 \% & \cellcolor{green!15}+6.5 \% \\
    \bottomrule
    \end{tabular}
    \end{adjustbox}
    \caption{Vocabulary-free fine-grained classification performance comparison among SOTA approaches. Metrics are denoted as cACC (clustering accuracy, \%) and sACC (semantic accuracy, \%). 
    For FiNDR and the approaches with automatically discovered vocabulary, the label discovery set includes 3 random unlabelled images per class, with the results averaged across 10 runs. 
    CLIP zero-shot results with known ground truth class names are for reference. Best results are bold, second-best are underlined.}
    \label{tab:main_results}
    \vspace{-6.0pt}
\end{table*}

\subsection{Experimental setup}

\subsubsection{Datasets.}
We evaluate our approach in the vocabulary-free regime on five widely popular fine-grained visual recognition benchmarks: CUB-200 (Birds-200) \cite{wah2011caltech}, Stanford Cars (Cars-196) \cite{hu2025car}, Stanford Dogs (Dogs-120) \cite{khosla2011novel}, Oxford Flowers (Flowers-102) \cite{nilsback2008automated}, and Oxford Pets (Pets-37) \cite{patino2016pets}.
Our evaluation protocol follows the vocabulary-free setup of \cite{finer}.
By default, we mimic a low-resource scenario by restricting the number of unlabelled images per class in the discovery set \( D_{\text{disc}} \) to only 3, randomly sampled from the discovery split.
Each test set \( D_{\text{test}} \) is unchanged and used for evaluation, with no overlap between discovery and test partitions: \( D_{\text{disc}} \cap D_{\text{test}} = \emptyset \).

\subsubsection{Evaluation Metrics.}
In the vocabulary-free setting, the set of labels produced by the model, \( \tilde{C^*} \), may not provide an exact one-to-one match with the ground-truth classes \( C \) \cite{CLIP, finer}.  To capture performance under this mismatch, we adopt two complementary metrics: Clustering Accuracy (cACC) and Semantic Accuracy (sACC).  
The goal of cACC is to evaluate how well the method assembles images that belong to the same true class, independent of whether the predicted textual label is correct. By contrast, sACC examines the semantics of those labels: a frozen language model scores the semantic relevance between each predicted name and its corresponding ground-truth category, so that a near-synonym incurs a smaller penalty than a semantically unrelated term.  
Used together, cACC verifies visual coherence while sACC ensures linguistic relevance, providing a balanced assessment. 

\subsubsection{Implementation Details.}
For the vocabulary discovery phase (Meta Information Generation and Class Names Prediction modules), we utilise the Qwen2.5-VL-72B-Instruct large multi-modal model (\cite{bai2025qwen25vltechnicalreport}).
For the Class Names Refinement module, as the vision-language model, a CLIP model \cite{radford2021clip,dosovitskiy2020image} based on ViT-L/14 visual encoder is used for filtration. In the vision-language Modalities Coupling block, we utilise a smaller ViT-B/16 base model for lower computational cost and faster inference.
For more details on the baselines and implementation setup, refer to Appendix \ref{app:baselines}, \ref{app:implement}, \ref{app:compute_cost}. 
Additionally, for prompt design and specifics, see Appendix \ref{app:prompts_details}.

\subsection{Comparison with State-Of-The-Art (SOTA)}  

Table \ref{tab:main_results} presents a detailed comparison of our FiNDR with various baselines across five fine-grained image classification benchmarks using both clustering accuracy (cACC) and semantic accuracy (sACC) metrics. 
Our approach is compared with competitive vocabulary-free baselines, including unsupervised clustering (KMeans \cite{krishna1999genetic}, Sinkhorn variants \cite{chen2022plot}), retrieval-based classification (WordNet \cite{WordNet}, CaSED \cite{conti2024vocabularyfreeimageclassification}), vision-language QA (BLIP-2) \cite{blip-2}, and current SOTA open-vocabulary methods like FineR \cite{finer} and E-FineR \cite{e-finer}.

The results indicate that FiNDR outperforms all existing methods, on average across all datasets, showcasing its strength in both clustering accuracy and semantic retrieval. Notably, our approach improves the average cACC by +9.5\% and sACC by +4.3\% compared to E-FineR, demonstrating its robustness in the vocabulary-free fine-grained recognition task.
The strength of FiNDR is particularly evident when we examine the results on datasets with subtle distinctions, such as CUB-200 and Dogs-120. Here, it significantly outperforms E-FineR, achieving +6.7\% in cACC and +6.8\% in sACC on CUB-200, and +6.3\% in cACC and +3.9\% in sACC on Dogs-120. These improvements are indicative of FiNDR’s ability to generate fine-grained labels that capture not only the visual similarities between instances but also their semantic relationships.
Specifically, the results suggest that FiNDR’s reasoning capabilities play a crucial role in its performance. Unlike previous SOTA methods that rely heavily on predefined vocabularies or multi-stage pipelines, our framework generates class labels dynamically based on visual features and direct reasoning about the objects in the image. This enables it not only to cluster similar images effectively, but also to assign more accurate, semantically grounded labels. 

Notably, on the Pets-37 dataset, FiNDR achieves an outstanding 86.5\% cACC and 83.7\% sACC, surpassing previous methods by significant relative margins of 18.7 \% and 9.9 \% respectively. 
The model's ability to provide specific breed names showcases its precision and its capability to distinguish breeds that appear visually similar but are distinct in terms of classification. 
This refinement is a direct consequence of FiNDR's reasoning-driven approach, which takes into account both visual cues and semantic meaning, leading to more accurate and grounded labels.

Interestingly, FiNDR shines on datasets like Flowers-102, where the improvement in cACC is particularly striking. Our method achieves a cACC of 79.8\%, which is a 15.0\% improvement over E-FineR, which highlights its ability to group flowers based on subtle visual features like petal shapes and colours. However, the modest improvement in sACC (+2.5\%) suggests that while FiNDR is excellent at clustering, it sometimes produces highly descriptive class names that, while visually accurate, may not always align with the biased ground-truth labels. This could be due to FiNDR’s ability to generate more detailed labels that go beyond the constraints of the official class names, which in some cases leads to a slight mismatch in semantic accuracy.


Overall, the results in Table \ref{tab:main_results} highlight that FiNDR consistently outperforms previous state-of-the-art methods in terms of clustering accuracy and semantic accuracy. This success stems from its unique approach of dynamically generating class labels through reasoning, without the reliance on predefined vocabularies or external knowledge bases. In this way, FiNDR's generalisation ability and robustness across diverse fine-grained datasets make it a strong contender in open-world visual recognition tasks.

\subsection{Analysis and Discussion}

\subsubsection{Vocabulary Quality.} 

To assess the quality of class names generated without a fixed vocabulary, we qualitatively compare predictions from our method, FiNDR, with the previous state-of-the-art, FineR, using examples from the Oxford Pets dataset (Figure \ref{fig:analysis_vocabulary}). The predicted labels are ranked by their similarity scores, and we specifically analyse the top-3 (highest similarity) and bottom-3 (lowest similarity) predictions from each method.
For the top-ranked predictions, FiNDR consistently produces accurate and semantically precise labels. 
While FineR sometimes generates partially correct labels, such as predicting ``American Cocker Spaniel" instead of the ``English Cocker Spaniel", FiNDR reliably identifies the exact breeds (e.g., correctly labelling ``Egyptian Mau" or ``Staffordshire Bull Terrier"). This accuracy highlights FiNDR’s ability to capture subtle visual distinctions and leverage deeper semantic cues, ensuring a more specific and correct identification compared to FineR.
The bottom-ranked predictions further underscore FiNDR’s robustness. 
FineR occasionally outputs semantically incorrect or overly generic predictions, exemplified by labels such as ``Possible Cat Breeds" or inaccurately identifying a dog with clearly white fur as a ``Golden Retriever". This can be explained by the FineR's multi-stage pipeline, which can propagate errors from previous stages and occasionally yield sub-optimal or incorrect labels.
In contrast, even FiNDR’s lower-confidence predictions remain semantically plausible and confined within their broad categories. FiNDR avoids severe semantic errors and placeholder labels, thanks to its integrated visual and taxonomic reasoning steps.

%
\begin{figure}[!t]
    \centering
    \includegraphics[width=0.89\columnwidth]{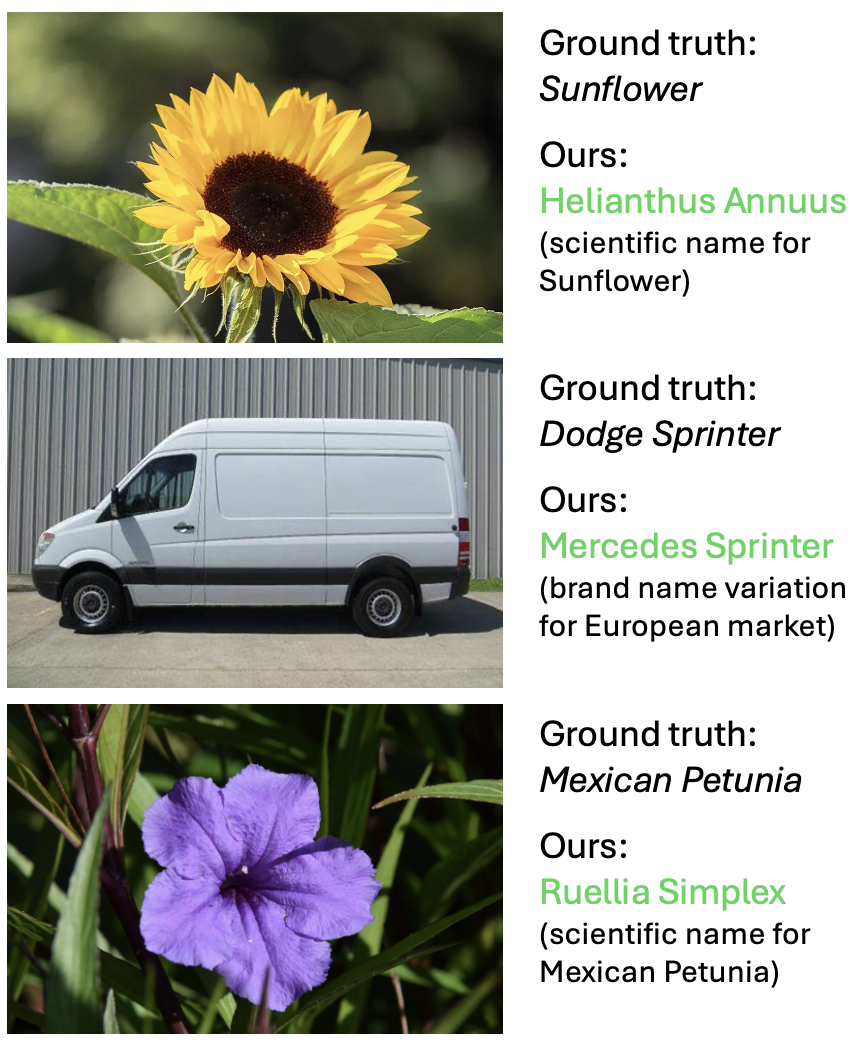} 
    \caption{Analysis and comparison of the class names generated by our FiNDR and the ground truth labels from Oxford Flowers and Stanford Cars datasets.
    It can be seen that our framework predicts correct label names, however, these names do not always fully match the biased, human-provided ground truth labels.
    This explains the higher increase in clustering accuracy, but a modest improvement in the semantic accuracy metric (also see Tab. \ref{tab:main_results}).
    %
    %
    }
    \label{fig:label_semantics}
    \vspace{-9.0pt}
\end{figure}

Overall, this analysis demonstrates FiNDR’s clear advantage in generating semantically precise, fine-grained labels. By effectively combining visual information with semantic reasoning and filtering mechanisms, FiNDR significantly improves the semantic validity and specificity of vocabulary-free predictions, surpassing FineR in both top- and bottom-ranked label quality.

\subsubsection{Label Semantics.} 
Our analysis reveals an intriguing discrepancy between clustering accuracy (cACC) and semantic accuracy (sACC) across certain datasets, notably Oxford Flowers and Stanford Cars. FiNDR significantly boosts clustering accuracy, effectively grouping visually similar images by accurately capturing subtle distinctions. However, improvements in semantic accuracy are comparatively modest. This pattern emerges primarily because FiNDR generates precise, detailed, and visually accurate labels that do not always align exactly with the single, rigid ground-truth label specified by the dataset.
Figure \ref{fig:label_semantics} clearly illustrates this phenomenon with specific examples. For instance, FiNDR correctly identifies a flower as ``Helianthus Annuus", the scientific name for ``Sunflower", whereas the dataset provides only the common name. Similarly, FiNDR precisely labels a vehicle as ``Mercedes Sprinter", a valid European market brand name variant, while the dataset’s ground-truth label is the North American market name ``Dodge Sprinter". 
Such discrepancies arise from the existence of multiple valid names for the same object, highlighting a limitation in current semantic evaluation protocols. The semantic accuracy metric currently credits only one predefined canonical label per class, thereby underestimating the true semantic quality and specificity of FiNDR’s predictions. This suggests the need for evaluation frameworks that not only account for a rigid list of fixed namings but also better accommodate the inherent diversity of potentially correct fine-grained labels.

\vspace{5.0pt}
\subsection{Ablation Study}


\vspace{3.0pt}
\subsubsection{Prompt Design.}
%
We conduct a thorough ablation study to analyse how different components of our prompt design influence vocabulary-free classification performance. Specifically, we compare three progressively enriched prompts: (i) the base prompt, which uses a straightforward, naive query (``What is the main object in the image?"), (ii) the meta information, which supplements the base prompt with additional meta-category information, and (iii) the expert prompt, which adds a contextual preamble such as ``You are an \{expert\} in the \{field\}" before the main query (see Appendix \ref{app:prompts_details}).

As shown in Table \ref{tab:ablation_prompt}, integrating meta-category information consistently enhances classification performance. Compared to the base prompt, the meta-enhanced prompt significantly increases clustering accuracy (cACC) by 1.1\%, 2.1\%, and 5.2\%, and semantic accuracy (sACC) by 2.2\%, 3.2\%, and 8.9\% for Birds, Cars, and Pets datasets, respectively. These improvements underscore the value of contextual semantic details in guiding the model toward more accurate predictions. For more analysis, refer to Appendix~\ref{app:extra_analysis}.
Adding expert notations further boosts semantic accuracy across all datasets, with increases of 0.3\% (Birds), 0.5\% (Cars), and 1.5\% (Pets), clearly demonstrating that explicitly positioning the model as a domain expert enhances semantic precision and interpretability. Although clustering accuracy experiences minor fluctuations (a slight 0.8\% decrease in Cars), the overall improvements in semantic accuracy justify the expert prompt strategy.

Overall, these results highlight the incremental and cumulative benefits of incorporating progressively richer contextual and semantic cues into prompt design, particularly emphasising the role of expert annotations in achieving refined semantic grounding in fine-grained classification tasks.
For more specifics on the prompt design and the exact utilised prompts, refer to Appendix \ref{app:prompts_details}.

\begin{table}[!t]
    \centering
    \normalsize
    \setlength{\tabcolsep}{1.5pt}
    \renewcommand{\arraystretch}{1.1}
    \begin{adjustbox}{max width=\textwidth}
    \begin{tabular}{ccc|cccccc}
    \toprule
    \multicolumn{3}{c|}{\textbf{Components}} & \multicolumn{6}{c}{\textbf{Accuracy}} \\
    \midrule
    \multicolumn{3}{c|}{Ours} & \multicolumn{2}{c|}{Birds} & \multicolumn{2}{c|}{Cars} & \multicolumn{2}{c}{Pets} \\
    \midrule
    Base & Meta & Expert & cACC & \multicolumn{1}{c|}{sACC} & cACC & \multicolumn{1}{c|}{sACC} & cACC & \multicolumn{1}{c}{sACC} \\
    \midrule
    \checkmark & $\times$ & $\times$ & 57.84 &  \multicolumn{1}{c|}{74.59} & 53.88 &  \multicolumn{1}{c|}{60.49} & 76.61 & 74.02 \\
    \checkmark & \checkmark & $\times$ & 58.90 & \multicolumn{1}{c|}{76.74} & \textbf{55.95} & \multicolumn{1}{c|}{63.64} & 81.82 & 82.96 \\ 
    %
    \checkmark & \checkmark & \checkmark & \textbf{59.33} & \multicolumn{1}{c|}{\textbf{77.04}} & 55.19 & \multicolumn{1}{c|}{\textbf{64.10}} & \textbf{84.13} & \textbf{84.46} \\
    \midrule
    \multicolumn{3}{c|}{$\Delta_{abs}$ \scriptsize{Ours to Base,\%}} & \cellcolor{green!15}+1.5 & \multicolumn{1}{c|}{\cellcolor{green!15}+2.5} & \cellcolor{green!15}+1.4 & \multicolumn{1}{c|}{\cellcolor{green!15}+3.7} & \cellcolor{green!15}+7.6 & \cellcolor{green!15}+10.5 \\
    \multicolumn{3}{c|}{$\Delta_{rel}$ \scriptsize{Ours to Base,\%}} & \cellcolor{green!15}+2.6 & \multicolumn{1}{c|}{\cellcolor{green!15}+3.3} & \cellcolor{green!15}+2.5 & \multicolumn{1}{c|}{\cellcolor{green!15}+6.0} & \cellcolor{green!15}+9.9 & \cellcolor{green!15}+14.2 \\    
    \bottomrule
    \end{tabular}
    \end{adjustbox}
    \caption{Ablation study for the components of our prompt design. 
    The performance is reported for the CUB-200, Stanford Dogs, and Oxford Pets datasets.
    Denoted components: base - the most basic and naive prompt (``What is the main object in the image?"), meta - including meta category information to the prompt, expert - pre-pending the expert's notation to the prompt.
    Best results 
    in bold.
    %
    %
    }    
    \label{tab:ablation_prompt}
\end{table}

\vspace{5.0pt}
\subsubsection{Explicit vs. Internal Reasoning.} 

We investigate two distinct reasoning strategies in fine-grained vocabulary-free classification: explicit reasoning and internal reasoning. Explicit reasoning involves prompt-guided step-by-step logical instructions explicitly articulated to the model, typically achieved via carefully engineered prompts (e.g., instructing the model to ``think step-by-step" before answering). In contrast, internal reasoning refers to implicit reasoning mechanisms integrated directly into advanced private models, allowing automatic inference without external prompting guidance. Currently, internal reasoning remains exclusive to proprietary LMMs, limiting its access and widespread adoption.

\begin{table}[!t]
    \centering
    \normalsize
    \setlength{\tabcolsep}{1.5pt}
    \renewcommand{\arraystretch}{1.1}
    \begin{adjustbox}{max width=\textwidth}
    \begin{tabular}{cc|c|c|c}
    \toprule
    \multicolumn{2}{c|}{\textbf{Birds-200}} & Qwen2.5-VL & \multicolumn{2}{c}{Gemini2.5-Flash} \\
    \midrule    
    \multicolumn{2}{c|}{Reasoning} & explicit & explicit & explicit + internal \\
    \midrule
    \multirow{2}{*}{Base} & cACC & 57.84 & 54.41 & 59.23 \\
     & sACC & 74.59 & 73.46 & 76.06 \\
    \midrule
    \multirow{2}{*}{Ours} & cACC & \underline{59.27} & 58.64 & \textbf{59.63} \\
     & sACC & 76.65 & \underline{76.86} & \textbf{78.04} \\
    \bottomrule
    \end{tabular}
    \end{adjustbox}
    \caption{Analysis of reasoning type benefits in a vocabulary-free setting on the Birds-200 dataset. 
    We consider open-sourced Qwen2.5-VL and closed-sourced Gemini2.5-Flash, in two reasoning variants: explicit (via prompt design) and internal (via the model's thinking configuration).
    Prompt options: Base - the most basic 
    prompt (``What is the main object in the image?"), Ours - our proposed prompt including meta category information and the expert's notation.
    %
    %
    Best results 
    in bold, second best 
    underlined.
    %
    %
    }    
    \label{tab:ablation_reasoning}
\end{table}

However, as demonstrated by our experiments on the CUB-200 dataset (Table \ref{tab:ablation_reasoning}), explicitly guided reasoning, combined with our 
prompting strategies, 
can significantly elevate the performance of public open-source models. For example, the open-source Qwen2.5-VL model \cite{bai2025qwen25vltechnicalreport} achieves 76.65\% semantic accuracy  using our carefully designed prompt, closely approaching the 78.04\% sACC obtained by the private Gemini2.5 model \cite{comanici2025gemini25pushingfrontier} employing internal reasoning. Moreover, this advanced explicit prompting improved clustering accuracy from 57.84\% (basic prompt) to 59.27\%, nearly matching the private model’s performance (59.63\% cACC). These findings underscore that sophisticated prompt engineering can effectively bridge the performance gap, making explicit reasoning a powerful and accessible tool for enhancing public LMMs in vocabulary-free fine-grained image classification tasks.

For additional analysis of the proposed system and components design, refer to Appendix \ref{app:extra_analysis}.

\section{Conclusion}

This work demonstrates that reasoning-augmented large multi-modal models can solve vocabulary-free fine-grained image classification end-to-end without relying on any human-curated label lists.
By leveraging an LMM to discover candidate names, verifying them with a vision-language model, and coupling textual and visual prototypes, our FiNDR framework forms a fully automated classifier that generalises across diverse domains. 
Extensive experiments on fine-grained benchmarks 
confirm that FiNDR not only establishes a new state-of-the-art margin over prior vocabulary-free methods but also surpasses zero-shot baselines which have access to ground-truth names. This observation overturns the long-held assumption that predefined vocabularies comprise an unreachable upper bound.
Ablation studies further demonstrate that explicit chain-of-thought prompting and expert-style instructions markedly boost naming quality and allow open-source LMMs to rival proprietary counterparts. 
Collectively, these findings position FiNDR as a scalable, interpretable, and resource-efficient foundation for open-world visual recognition and highlight the broader promise of integrating structured reasoning with multi-modal foundation models.

{
    \small
    \bibliographystyle{ieeenat_fullname}
    \bibliography{main}

\begin{thebibliography}{54}
\providecommand{\natexlab}[1]{#1}
\providecommand{\url}[1]{\texttt{#1}}
\expandafter\ifx\csname urlstyle\endcsname\relax
  \providecommand{\doi}[1]{doi: #1}\else
  \providecommand{\doi}{doi: \begingroup \urlstyle{rm}\Url}\fi

\bibitem[Akata et~al.(2015)Akata, Reed, Walter, Lee, and Schiele]{akata2015evaluation}
Zeynep Akata, Scott Reed, Daniel Walter, Honglak Lee, and Bernt Schiele.
\newblock Evaluation of output embeddings for fine-grained image classification.
\newblock In \emph{CVPR}, 2015.

\bibitem[Bai et~al.(2024)Bai, Xu, Zhou, et~al.]{bai2024internvl2}
Jing Bai, Chen Xu, Yichao Zhou, et~al.
\newblock Internvl 2.0: Scaling up vision foundation models and aligning for multimodal understanding.
\newblock \emph{arXiv preprint arXiv:2404.XXXX}, 2024.

\bibitem[Bai et~al.(2025)Bai, Chen, Liu, Wang, Ge, Song, Dang, Wang, Wang, Tang, Zhong, Zhu, Yang, Li, Wan, Wang, Ding, Fu, Xu, Ye, Zhang, Xie, Cheng, Zhang, Yang, Xu, and Lin]{bai2025qwen25vltechnicalreport}
Shuai Bai, Keqin Chen, Xuejing Liu, Jialin Wang, Wenbin Ge, Sibo Song, Kai Dang, Peng Wang, Shijie Wang, Jun Tang, Humen Zhong, Yuanzhi Zhu, Mingkun Yang, Zhaohai Li, Jianqiang Wan, Pengfei Wang, Wei Ding, Zheren Fu, Yiheng Xu, Jiabo Ye, Xi Zhang, Tianbao Xie, Zesen Cheng, Hang Zhang, Zhibo Yang, Haiyang Xu, and Junyang Lin.
\newblock Qwen2.5-vl technical report, 2025.

\bibitem[Branson et~al.(2014)Branson, Van~Horn, Belongie, and Perona]{branson2014bird}
Steve Branson, Grant Van~Horn, Serge Belongie, and Pietro Perona.
\newblock Bird species categorization using pose normalized deep convolutional nets.
\newblock \emph{arXiv preprint arXiv:1406.2952}, 2014.

\bibitem[Caron et~al.(2021)Caron, Touvron, Misra, J{\'e}gou, Mairal, Bojanowski, and Joulin]{caron2021dino}
Mathilde Caron, Hugo Touvron, Ishan Misra, Herv{\'e} J{\'e}gou, Julien Mairal, Piotr Bojanowski, and Armand Joulin.
\newblock Emerging properties in self-supervised vision transformers.
\newblock In \emph{Proceedings of the IEEE/CVF International Conference on Computer Vision (ICCV)}, 2021.

\bibitem[Chang et~al.(2020)Chang, Ding, Xie, Bhunia, Li, Ma, Wu, Guo, and Song]{chang2020devil}
Dongliang Chang, Yifeng Ding, Jiyang Xie, Ayan~Kumar Bhunia, Xiaoxu Li, Zhanyu Ma, Ming Wu, Jun Guo, and Yi-Zhe Song.
\newblock The devil is in the channels: Mutual-channel loss for fine-grained image classification.
\newblock \emph{IEEE Transactions on Image Processing}, 29:\penalty0 4683--4695, 2020.

\bibitem[Chen et~al.(2022)Chen, Yao, Song, Li, Rao, and Zhang]{chen2022plot}
Guangyi Chen, Weiran Yao, Xiangchen Song, Xinyue Li, Yongming Rao, and Kun Zhang.
\newblock Plot: Prompt learning with optimal transport for vision-language models.
\newblock \emph{arXiv preprint arXiv:2210.01253}, 2022.

\bibitem[Chou et~al.(2022)Chou, Lin, and Kao]{chou2022novel}
Po-Yung Chou, Cheng-Hung Lin, and Wen-Chung Kao.
\newblock A novel plug-in module for fine-grained visual classification.
\newblock \emph{arXiv preprint arXiv:2202.03822}, 2022.

\bibitem[Choudhury et~al.(2021)Choudhury, Laina, Rupprecht, and Vedaldi]{DBLP:journals/corr/abs-2111-03651}
Subhabrata Choudhury, Iro Laina, Christian Rupprecht, and Andrea Vedaldi.
\newblock The curious layperson: Fine-grained image recognition without expert labels.
\newblock \emph{CoRR}, abs/2111.03651, 2021.

\bibitem[Conti et~al.(2024)Conti, Fini, Mancini, Rota, Wang, and Ricci]{conti2024vocabularyfreeimageclassification}
Alessandro Conti, Enrico Fini, Massimiliano Mancini, Paolo Rota, Yiming Wang, and Elisa Ricci.
\newblock Vocabulary-free image classification, 2024.

\bibitem[Dai et~al.(2023)Dai, Yu, Jin, and et~al.]{dai2023instructblip}
Wenliang Dai, Zhe Yu, Lijuan Jin, and et al.
\newblock Instructblip: Towards general-purpose vision-language models with instruction tuning.
\newblock In \emph{Advances in Neural Information Processing Systems (NeurIPS)}, 2023.

\bibitem[Demidov et~al.(2024)Demidov, Shtanchaev, Mihaylov, and Almansoori]{Demidov_2024_BMVC}
Dmitry Demidov, Abduragim Shtanchaev, Mihail~Minkov Mihaylov, and Mohammad Almansoori.
\newblock Extract more from less: Efficient fine-grained visual recognition in low-data regimes.
\newblock In \emph{35th British Machine Vision Conference 2024, {BMVC} 2024, Glasgow, UK, November 25-28, 2024}. BMVA, 2024.

\bibitem[Demidov et~al.(2025)Demidov, Zaheer, Thawakar, Khan, and Khan]{e-finer}
Dmitry Demidov, Zaigham Zaheer, Omkar Thawakar, Salman Khan, and Fahad~Shahbaz Khan.
\newblock Vocabulary-free fine-grained visual recognition via enriched contextually grounded vision-language model, 2025.

\bibitem[Dosovitskiy et~al.(2020)Dosovitskiy, Beyer, Kolesnikov, Weissenborn, Zhai, Unterthiner, Dehghani, Minderer, Heigold, Gelly, et~al.]{dosovitskiy2020image}
Alexey Dosovitskiy, Lucas Beyer, Alexander Kolesnikov, Dirk Weissenborn, Xiaohua Zhai, Thomas Unterthiner, Mostafa Dehghani, Matthias Minderer, Georg Heigold, Sylvain Gelly, et~al.
\newblock An image is worth 16x16 words: Transformers for image recognition at scale.
\newblock \emph{arXiv preprint arXiv:2010.11929}, 2020.

\bibitem[Gao et~al.(2016)Gao, Beijbom, Zhang, and Darrell]{7780410}
Yang Gao, Oscar Beijbom, Ning Zhang, and Trevor Darrell.
\newblock Compact bilinear pooling.
\newblock In \emph{2016 IEEE Conference on Computer Vision and Pattern Recognition (CVPR)}, pages 317--326, 2016.

\bibitem[Gheorghe~Comanici(2025)]{comanici2025gemini25pushingfrontier}
et~al. Gheorghe~Comanici.
\newblock Gemini 2.5: Pushing the frontier with advanced reasoning, multimodality, long context, and next generation agentic capabilities, 2025.

\bibitem[Han et~al.(2024)Han, Huang, Li, Vaze, Li, and Jia]{han2024whatsnameclassindices}
Kai Han, Xiaohu Huang, Yandong Li, Sagar Vaze, Jie Li, and Xuhui Jia.
\newblock What's in a name? beyond class indices for image recognition, 2024.

\bibitem[Han et~al.(2020)Han, Fu, and Yang]{han2020learning}
Zongyan Han, Zhenyong Fu, and Jian Yang.
\newblock Learning the redundancy-free features for generalized zero-shot object recognition.
\newblock In \emph{CVPR}, 2020.

\bibitem[Han et~al.(2021)Han, Fu, Chen, and Yang]{han2021contrastive}
Zongyan Han, Zhenyong Fu, Shuo Chen, and Jian Yang.
\newblock Contrastive embedding for generalized zero-shot learning.
\newblock In \emph{CVPR}, 2021.

\bibitem[Hu et~al.(2025)Hu, Li, Yan, Shao, and Luo]{hu2025car}
Yutao Hu, Sen Li, Jincheng Yan, Wenqi Shao, and Xiaoyan Luo.
\newblock Car-1000: A new large scale fine-grained visual categorization dataset.
\newblock \emph{arXiv preprint arXiv:2503.12385}, 2025.

\bibitem[Jinguo~Zhu(2024)]{zhu2025internvl3exploringadvancedtraining}
et~al. Jinguo~Zhu.
\newblock Internvl3: Exploring advanced training and test-time recipes for open-source multimodal models, 2024.

\bibitem[Khosla et~al.(2011)Khosla, Jayadevaprakash, Yao, and Li]{khosla2011novel}
Aditya Khosla, Nityananda Jayadevaprakash, Bangpeng Yao, and Fei-Fei Li.
\newblock Novel dataset for fine-grained image categorization: Stanford dogs.
\newblock In \emph{Proc. CVPR workshop on fine-grained visual categorization (FGVC)}, 2011.

\bibitem[Kojima et~al.(2022)Kojima, Gu, Reid, Matsuo, and Iwasawa]{kojima2022zero_shot_reasoners}
Takeshi Kojima, Shixiang~Shane Gu, Machel Reid, Yutaka Matsuo, and Yusuke Iwasawa.
\newblock Large language models are zero-shot reasoners.
\newblock In \emph{Advances in Neural Information Processing Systems (NeurIPS)}, 2022.

\bibitem[Krishna and Murty(1999)]{krishna1999genetic}
K Krishna and M~Narasimha Murty.
\newblock Genetic k-means algorithm.
\newblock \emph{IEEE Transactions on Systems, Man, and Cybernetics, Part B (Cybernetics)}, 29\penalty0 (3):\penalty0 433--439, 1999.

\bibitem[Lagunas et~al.(2023)Lagunas, Impata, Martinez, Fernandez, Georgakis, Braun, and Bertrand]{lagunas2023transfer}
Manuel Lagunas, Brayan Impata, Victor Martinez, Virginia Fernandez, Christos Georgakis, Sofia Braun, and Felipe Bertrand.
\newblock Transfer learning for fine-grained classification using semi-supervised learning and visual transformers.
\newblock \emph{arXiv preprint arXiv:2305.10018}, 2023.

\bibitem[Lamott and Shakir(2024)]{lamott2024leveraging}
Marcel Lamott and Muhammad~Armaghan Shakir.
\newblock Leveraging distillation techniques for document understanding: A case study with flan-t5.
\newblock \emph{arXiv preprint arXiv:2409.11282}, 2024.

\bibitem[Li et~al.(2023{\natexlab{a}})Li, Li, Savarese, and Hoi]{blip-2}
Junnan Li, Dongxu Li, Silvio Savarese, and Steven Hoi.
\newblock Blip-2: Bootstrapping language-image pre-training with frozen image encoders and large language models, 2023{\natexlab{a}}.

\bibitem[Li et~al.(2023{\natexlab{b}})Li, Li, Xie, and et~al.]{li2023blip2}
Jiasen Li, Dongxu Li, Yixuan Xie, and et al.
\newblock Blip-2: Bootstrapping language-image pre-training with frozen image encoders and large language models.
\newblock In \emph{Proceedings of the 40th International Conference on Machine Learning (ICML)}, 2023{\natexlab{b}}.

\bibitem[Lin et~al.(2015)Lin, RoyChowdhury, and Maji]{7410527}
Tsung-Yu Lin, Aruni RoyChowdhury, and Subhransu Maji.
\newblock Bilinear cnn models for fine-grained visual recognition.
\newblock In \emph{2015 IEEE International Conference on Computer Vision (ICCV)}, pages 1449--1457, 2015.

\bibitem[Liu et~al.(2023)Liu, Li, Wu, and Lee]{liu2023llava}
Haotian Liu, Chunyuan Li, Qingyang Wu, and Yong~Jae Lee.
\newblock Visual instruction tuning.
\newblock In \emph{Advances in Neural Information Processing Systems (NeurIPS) Datasets and Benchmarks Track}, 2023.

\bibitem[Liu et~al.(2024)Liu, Roy, Li, Zhong, Sebe, and Ricci]{finer}
Mingxuan Liu, Subhankar Roy, Wenjing Li, Zhun Zhong, Nicu Sebe, and Elisa Ricci.
\newblock Democratizing fine-grained visual recognition with large language models, 2024.

\bibitem[Miller(1995)]{WordNet}
George~A. Miller.
\newblock Wordnet: a lexical database for english.
\newblock \emph{Commun. ACM}, 38\penalty0 (11):\penalty0 39–41, 1995.

\bibitem[Minderer et~al.(2022)Minderer, Gritsenko, Sun, and et~al.]{minderer2022simple}
Matthias Minderer, Alexey Gritsenko, Chen Sun, and et al.
\newblock Simple open-vocabulary object detection with vision transformers.
\newblock In \emph{Proceedings of the European Conference on Computer Vision (ECCV)}, 2022.

\bibitem[Nilsback and Zisserman(2008)]{nilsback2008automated}
Maria-Elena Nilsback and Andrew Zisserman.
\newblock Automated flower classification over a large number of classes.
\newblock In \emph{2008 Sixth Indian conference on computer vision, graphics \& image processing}, pages 722--729. IEEE, 2008.

\bibitem[OpenAI(2024)]{openai2024gpt4technicalreport}
et~al. OpenAI.
\newblock Gpt-4 technical report, 2024.

\bibitem[Parikh and Grauman(2011)]{parikh2011relative}
Devi Parikh and Kristen Grauman.
\newblock Relative attributes.
\newblock In \emph{ICCV}, 2011.

\bibitem[Patino et~al.(2016)Patino, Cane, Vallee, and Ferryman]{patino2016pets}
Luis Patino, Tom Cane, Alain Vallee, and James Ferryman.
\newblock Pets 2016: Dataset and challenge.
\newblock In \emph{Proceedings of the IEEE conference on computer vision and pattern recognition workshops}, pages 1--8, 2016.

\bibitem[Radford et~al.(2021{\natexlab{a}})Radford, Kim, Hallacy, Ramesh, Goh, Agarwal, Sastry, Askell, Mishkin, Clark, Krueger, and Sutskever]{CLIP}
Alec Radford, Jong~Wook Kim, Chris Hallacy, Aditya Ramesh, Gabriel Goh, Sandhini Agarwal, Girish Sastry, Amanda Askell, Pamela Mishkin, Jack Clark, Gretchen Krueger, and Ilya Sutskever.
\newblock Learning transferable visual models from natural language supervision, 2021{\natexlab{a}}.

\bibitem[Radford et~al.(2021{\natexlab{b}})Radford, Kim, Hallacy, Ramesh, Goh, Agarwal, Sastry, Askell, Mishkin, Clark, et~al.]{radford2021clip}
Alec Radford, Jong~Wook Kim, Chris Hallacy, Aditya Ramesh, Gabriel Goh, Sandhini Agarwal, Girish Sastry, Amanda Askell, Pamela Mishkin, Jack Clark, et~al.
\newblock Learning transferable visual models from natural language supervision.
\newblock In \emph{Proceedings of the 38th International Conference on Machine Learning (ICML)}, 2021{\natexlab{b}}.

\bibitem[Schmarje et~al.(2021)Schmarje, Santarossa, Schr{\"o}der, and Koch]{schmarje2021survey}
Lars Schmarje, Monty Santarossa, Simon-Martin Schr{\"o}der, and Reinhard Koch.
\newblock A survey on semi-, self-and unsupervised learning for image classification.
\newblock \emph{IEEE Access}, 9:\penalty0 82146--82168, 2021.

\bibitem[Team(2024)]{qwen2vl2024}
Qwen-VL Team.
\newblock Qwen2-vl: Enhancing vision-language understanding with better alignment and vision tokenization.
\newblock \emph{arXiv preprint arXiv:2407.XXXX}, 2024.

\bibitem[Vaze et~al.(2022)Vaze, Han, Vedaldi, and Zisserman]{vaze2022gcd}
Sagar Vaze, Kai Han, Andrea Vedaldi, and Andrew Zisserman.
\newblock Generalized category discovery.
\newblock In \emph{Proceedings of the IEEE/CVF Conference on Computer Vision and Pattern Recognition (CVPR)}, 2022.

\bibitem[Wah et~al.(2011)Wah, Branson, Welinder, Perona, and Belongie]{wah2011caltech}
Catherine Wah, Steve Branson, Peter Welinder, Pietro Perona, and Serge Belongie.
\newblock The caltech-ucsd birds-200-2011 dataset.
\newblock 2011.

\bibitem[Wang et~al.(2021)Wang, Yu, and Gao]{Wang2021FeatureFV}
Jun Wang, Xiaohan Yu, and Yongsheng Gao.
\newblock Feature fusion vision transformer for fine-grained visual categorization.
\newblock In \emph{British Machine Vision Conference}, 2021.

\bibitem[Wang et~al.(2022)Wang, Sun, Yang, and Yang]{V2L}
Wenhao Wang, Yifan Sun, Zongxin Yang, and Yi Yang.
\newblock V$^2$l: Leveraging vision and vision-language models into large-scale product retrieval, 2022.

\bibitem[Wang et~al.(2023{\natexlab{a}})Wang, Sun, Li, and Yang]{TransHP}
Wenhao Wang, Yifan Sun, Wei Li, and Yi Yang.
\newblock Transhp: Image classification with hierarchical prompting, 2023{\natexlab{a}}.

\bibitem[Wang et~al.(2023{\natexlab{b}})Wang, Wei, Schuurmans, et~al.]{wang2023self_consistency}
Xuezhi Wang, Jason Wei, Dale Schuurmans, et~al.
\newblock Self-consistency improves chain of thought reasoning in language models.
\newblock In \emph{International Conference on Learning Representations (ICLR)}, 2023{\natexlab{b}}.

\bibitem[Wei et~al.(2022)Wei, Wang, Schuurmans, et~al.]{wei2022cot}
Jason Wei, Xuezhi Wang, Dale Schuurmans, et~al.
\newblock Chain-of-thought prompting elicits reasoning in large language models.
\newblock In \emph{Advances in Neural Information Processing Systems (NeurIPS)}, 2022.

\bibitem[Xu et~al.(2021)Xu, Zhang, Hu, Wang, Wang, Wei, Bai, and Liu]{xu2021end}
Mengde Xu, Zheng Zhang, Han Hu, Jianfeng Wang, Lijuan Wang, Fangyun Wei, Xiang Bai, and Zicheng Liu.
\newblock End-to-end semi-supervised object detection with soft teacher.
\newblock In \emph{Proceedings of the IEEE/CVF International Conference on Computer Vision}, pages 3060--3069, 2021.

\bibitem[Zhang et~al.(2014)Zhang, Donahue, Girshick, and Darrell]{zhang2014part}
Ning Zhang, Jeff Donahue, Ross Girshick, and Trevor Darrell.
\newblock Part-based r-cnns for fine-grained category detection.
\newblock In \emph{Computer Vision--ECCV 2014: 13th European Conference, Zurich, Switzerland, September 6-12, 2014, Proceedings, Part I 13}, pages 834--849. Springer, 2014.

\bibitem[Zhang et~al.(2023)Zhang, Li, Wang, and Wang]{zhang2023cased}
Ruohao Zhang, Yixuan Li, Yuting Wang, and Yu-Xiong Wang.
\newblock Cased: Category-agnostic semantic discovery for open-world recognition.
\newblock In \emph{Proceedings of the IEEE/CVF International Conference on Computer Vision (ICCV)}, 2023.

\bibitem[Zheng et~al.(2019)Zheng, Fu, Zha, and Luo]{10.5555/3454287.3454672}
Heliang Zheng, Jianlong Fu, Zheng-Jun Zha, and Jiebo Luo.
\newblock \emph{Learning deep bilinear transformation for fine-grained image representation}.
\newblock Curran Associates Inc., Red Hook, NY, USA, 2019.

\bibitem[Zhou et~al.(2025)Zhou, Tu, Wang, Wang, Muennighoff, Nie, Choi, Zou, Deng, Yan, Fan, Xie, Yao, and Ye]{zhou2025visualizingstepreasoningmira}
Yiyang Zhou, Haoqin Tu, Zijun Wang, Zeyu Wang, Niklas Muennighoff, Fan Nie, Yejin Choi, James Zou, Chaorui Deng, Shen Yan, Haoqi Fan, Cihang Xie, Huaxiu Yao, and Qinghao Ye.
\newblock When visualizing is the first step to reasoning: Mira, a benchmark for visual chain-of-thought, 2025.

\bibitem[Zhuang et~al.(2020)Zhuang, Wang, and Qiao]{zhuang2020learning}
Peiqin Zhuang, Yali Wang, and Yu Qiao.
\newblock Learning attentive pairwise interaction for fine-grained classification.
\newblock In \emph{Proceedings of the AAAI conference on artificial intelligence}, pages 13130--13137, 2020.

\end{thebibliography}
}
\clearpage

\appendix
\setcounter{page}{1}
\maketitlesupplementary

\section{Appendix}

\subsection{Additional Experimental Details}

\subsubsection{Baselines.}
\label{app:baselines}

The fine-grained visual recognition task approached without any expert annotations is currently an under-explored domain, with a few reference methods existing.
To offer a broader comparison and evaluation, we consider the strong baselines detailed below:
\begin{itemize}
    \item [(i)] \textbf{CLIP Zero-Shot:} the pre-defined and static ground-truth class names are supplied to CLIP as text prompts, mirroring expert knowledge and yielding an upper-bound performance estimate.  

    \item [(ii)] \textbf{WordNet Baseline:} CLIP is queried with a large lexicon of 119{,}000 nouns drawn from WordNet \cite{WordNet}.  
    
    \item [(iii)] \textbf{BLIP-2} \cite{blip-2} \textbf{and Flan-T5xxl} \cite{lamott2024leveraging}: a VQA-style approach that prompts the model to answer “\emph{What is the name of the main object in this image?}” in order to identify the primary object.  
    
    \item [(iv)] \textbf{SCD} \cite{han2024whatsnameclassindices}: images are first clustered, after which candidate labels are refined with CLIP using a combined vocabulary from WordNet and Wikipedia bird names. 
    
    \item [(v)] \textbf{CaSED} \cite{conti2024vocabularyfreeimageclassification}: captions retrieved from a large-scale knowledge base are parsed, and nouns are filtered with CLIP to extract class names.  
    
    \item [(vi)] \textbf{KMeans} clustering applied to CLIP visual embeddings \cite{krishna1999genetic}.  
    
    \item [(vii)] \textbf{Sinkhorn–Knopp Clustering:} a parametric method run on features from CLIP or DINO.  
    
    \item [(viii)] \textbf{FineR-based} methods, including FineR \cite{finer} and E-FineR \cite{e-finer}.
    These methods employ a dynamic vocabulary discovery pipeline. Specifically, they utilise a small set of unlabelled images to discover the potential class names using an LLM (such as ChatGPT-4 \cite{openai2024gpt4technicalreport}) and a VLM (such as CLIP \cite{radford2021clip}).
    In this approach the the recognition process is decoupled into multiple independent stages involving different models, introducing potential gaps and inconsistencies. Moreover, the predictions at intermediate steps are often not image-specific due to text-only processing. This setup may be prone to error accumulation and mispredictions, making FineR-based methods less reliable, especially in the presence of high image variance and intra-class diversity.
\end{itemize}

All baselines are evaluated with the CLIP ViT-B/16 vision encoder \cite{radford2021clip,dosovitskiy2020image}.

\begin{figure}[!t]
\centering
  \begin{mybox}{grey}{1}
  \textbf{Input}:
    \justifying
    \scriptsize{
    \begin{verbatim}
You are given a set of images representing a 
specific object category. Analyze these images 
and provide information about the main object 
in the images:                                                               
1. The category describing these specific objects 
(sungular and plural forms).                                  
2. The word typically used to describe a unit 
(or a sub-category) of this category, to distinct 
such specific similar objects (singular and plural 
forms).
3. The word typically used to describe a 
recognised expert or professional who studied 
this category and is able to easily distinct 
its units.

Please provide this information in this specific 
format as a JSON object with the following fields:
{
    "category_singular": "<category_singular>",
    "category_plural": "<category_plural>",
    "unit_singular": "<unit_singular>",
    "unit_plural": "<unit_plural>",
    "expert_name": "<expert_name>"
}

Do not provide any additional word or information.
    \end{verbatim}
    }
    \vspace{-1.5em}
  \end{mybox}
    \vspace{-1.5em}
  \begin{mybox}{white}{1}
  \textbf{Output}:
      \scriptsize{
      \begin{verbatim}
{
    "category_singular": "bird",
    "category_plural": "birds",
    "unit_singular": "species",
    "unit_plural": "species",
    "expert_name": "ornithologist"
}
    \end{verbatim}
    }
  \end{mybox}
  \caption{Example of the meta prompt input and output for the CUB-200 dataset.}
  \label{fig:prompt_meta}
\end{figure}

\subsubsection{Implementation Details.}
\label{app:implement}

For the vocabulary discovery phase (Meta Information Generation and Class Names Prediction modules), we utilise the Qwen2.5-VL-72B-Instruct large multi-modal model (\cite{bai2025qwen25vltechnicalreport}) accessed through the public API services on the OpenRouter platform. 
Specifically, the chosen model is initialised with the pre-trained weights from \cite{bai2025qwen25vltechnicalreport} and has a maximum context length of 33000 tokens and a maximum output of 33000 tokens as well.
The default recommended generation hyperparameters are kept unchanged.
Specifics about the used prompt are available in App. \ref{app:prompts_details}.

For the Class Names Refinement module, as the vision-language model, a CLIP model \cite{radford2021clip,dosovitskiy2020image} based on ViT-L/14 visual encoder is used for filtration, while in the vision-language Modalities Coupling block, we utilise a smaller ViT-B/16 base model for faster inference.
The hyper-parameters for vision-language fusion is set to \( \alpha = 0.7 \) (for details, refer to \ref{app:alpha_ablation}), and the number of applied data augmentations is set to \( K = 10 \).

\vspace{+1.0em}
\subsection{Prompt Design}
\label{app:prompts_details}

For every image \(x_i\) in the discovery split \(\mathcal{D}_{\mathrm{disc}}\), the large multi-modal model is utilised in two consecutive calls: to first predict meta-information about the object in the image and then a candidate class name for this object.
We first obtain dataset‑level meta‑information by querying the model with a small, randomly sampled context set of three images.
The request for meta-information \(m^{\star}\), queries the LMM for
\textit{(i)} a broad taxonomic group
\textit{(ii)} the granularity unit within that group, and  
\textit{(iii)} a domain specialist name, needed for fine discrimination. 
Refer to Figure \ref{fig:prompt_meta} for a specific example of this prompt.

Given this frozen context, each individual image \(x_i\) is then passed, together with \(m^{\star}\), to the same backbone model under a second prompt that requests for the unique fine‑grained label \(\tilde{c}_i\). Refer to Figure \ref{fig:prompt_main} for a specific example of this prompt.

\begin{figure}[h]
\centering
  \vspace{+1.0em}
  \begin{mybox}{grey}{1}
  \textbf{Input}:
    \scriptsize{
    \begin{verbatim}
<image: Nighthawk_Nighthawk_0028_82636.jpg>

You are a professional ornithologist and an expert 
in bird classification.
     
What is the exact bird species in the provided
image?
    \end{verbatim}
    }
    \vspace{-1.5em}
  \end{mybox}
    \vspace{-1.5em}
  \begin{mybox}{white}{1}
  \textbf{Output}:
      \scriptsize{
      \begin{verbatim}
The bird in the image is a **Common Nighthawk 
(Chordeiles minor)**.

### Key Features:
- **Shape and Size:** This bird has a slender, 
streamlined body and long, pointed wings, which 
are adapted for agile, rapid flight.
- **Plumage:** The Common Nighthawk has cryptic 
brownish-gray plumage with white spots near the 
wingtips, a white bar across the wing, and darker 
underparts.
- **Flight Progression:** Common Nighthawks are 
often seen flying high and moving erratically as 
they chase insects, a behavior captured in this 
image.

This species is part of the *Caprimulgidae* family, 
also known as nightjars, and is nocturnal or 
crepuscular in its habits.
    \end{verbatim}
    }
  \end{mybox}
  \caption{Example of the main prompt input and output for the CUB-200 dataset.}
  \label{fig:prompt_main}
\end{figure}

Finally, 
every raw prediction \(\tilde{c}_i\) is standardised to discard unusable outputs.  
This post‑processing normalises whitespace, capitalisation, and pluralisation. Additionally, it filters out syntactically corrupted strings and rejects names that remain overly generic after normalisation.
Refer to Figure \ref{fig:prompt_service} for the specific example of this prompt.

\begin{figure}[h]
\centering
  \vspace{+1.0em}
  \begin{mybox}{grey}{1}
  \textbf{Input}:
    \scriptsize{
    \begin{verbatim}
Convert the below text containing suggested bird 
species to a Python dictionary object, where a 
key is an index and the value is a suggestion of 
the specific bird species. 
Only use the final bird species prediction(s), 
do not use any intermediate suggestions.
Remove duplicated suggestions and unsepcific bird 
species. 
Also keep the numbered order of the suggestions 
with 1 as a starting index.
Make sure to only use English letters. 
Add a space between seprate words if not done 
in the suggested bird species and capitalize 
abbreviations and first letters of normal words.
    \end{verbatim}
    }
    \vspace{-1.5em}
  \end{mybox}
    \vspace{-1.5em}
  \begin{mybox}{white}{1}
  \textbf{Output}:
      \scriptsize{
      \begin{verbatim}
{"1": "Common Nighthawk"} 
    \end{verbatim}
    }
  \end{mybox}
  \caption{Example of the service prompt input and output for the CUB-200 dataset.}
  \label{fig:prompt_service}
\end{figure}

\subsection{Additional Related Work}
\label{app:extra_related_work}

\subsubsection{Fine-grained Visual Recognition.}

Fine-grained visual recognition (FGVR) targets the classification of very similar subordinate classes that lie within a broader super-category (for instance, bird species or car models). Because such classes differ only subtly, traditional FGVR pipelines rely on labelled images and auxiliary expert annotations. Existing methods can be grouped into two main camps: (i) feature-encoding approaches, which improve recognition through richer feature representations (e.g., bilinear pooling); and (ii) localisation-based approaches, which first locate and then concentrate on the most discriminative regions of the image.  

Early work employed pre-trained R-CNNs \cite{zhang2014part} and part detectors \cite{branson2014bird} to align corresponding object parts across images. More recent solutions favour end-to-end architectures \cite{zhuang2020learning} in which multiple backbones produce a shared feature vector that is compared against instance-specific representations, enabling the model to separate challenging categories \cite{7410527, 7780410, 10.5555/3454287.3454672}.  

A complementary line of research develops robust loss functions \cite{chang2020devil,Demidov_2024_BMVC}. Typical designs include a \textit{discriminability} term that encourages every channel linked to a class to be informative, often via channel-wise attention, and a \textit{diversity} term that makes those channels mutually exclusive. Related methods likewise focus on stronger features, for example by aggregating salient tokens from each transformer layer to recover local and mid-level cues \cite{Wang2021FeatureFV}.  

Some recent studies \cite{lagunas2023transfer, chou2022novel} leverage the large Vision Transformer (ViT) backbone \cite{dosovitskiy2020image} in semi-supervised settings to boost performance, though these models demand expensive pre-training and tend to be slower at inference.  

The latest techniques, such as TransHP \cite{TransHP} and V2L \cite{V2L}, pair vision-language modelling with learned soft prompts to enhance fine-grained discrimination. Nonetheless, they still assume the availability of predefined fine-grained labels.

\subsection{Additional Analysis}
\label{app:extra_analysis}

\subsubsection{Public vs. Private LMMs.} 

Recent progress has enabled open-source LMMs to achieve performance on par with previously dominant proprietary models. We compare Qwen2.5-VL \cite{bai2025qwen25vltechnicalreport} (a public open-source LMM) with Gemini2.5-Flash \cite{comanici2025gemini25pushingfrontier} (a private closed-source LMM) as representative examples.
The results are summarised in Table \ref{tab:public_vs_private}, where both models show comparable performance. 
Specifically, with our advanced prompt design, the two models obtain nearly identical average clustering accuracy (cACC: 67.94 \% for Qwen vs.\ 67.32 \% for Gemini), while Qwen achieves a substantially higher semantic accuracy (sACC: 70.51 \% vs.\ 63.30 \%).  This result indicates that an open-source LMM, when provided explicit reasoning instructions, can leverage its latent knowledge as effectively as a closed-source counterpart.

\begin{table}[!h]
    \centering
    \normalsize
    \setlength{\tabcolsep}{5.3pt}
    \renewcommand{\arraystretch}{1.1}
    \begin{adjustbox}{max width=\textwidth}
    \begin{tabular}{cc||c|c||c|c}
    \toprule
    \multicolumn{2}{c||}{\textbf{Dataset}} & \multicolumn{2}{c||}{Qwen2.5-VL} & \multicolumn{2}{c}{Gemini2.5-Flash}\\
    \midrule    
    \multicolumn{2}{c||}{Dataset-specifics} & $\times$ & \checkmark & $\times$ & \checkmark \\
    \midrule
    \midrule 
    \multirow{2}{*}{Birds} & cACC & 58.88 & 58.89 & 60.81 & 59.20 \\
     & sACC & 76.91 & 76.81 & 65.37 & 77.27 \\
    \midrule
    \multirow{2}{*}{Cars} & cACC & 55.94 & 54.46 & 55.01 & 54.61 \\
     & sACC & 64.37 & 65.32 & 58.84 & 63.04 \\
    \midrule
    \multirow{2}{*}{Dogs} & cACC & 58.09 & 57.94 & 56.11 & 58.42 \\
     & sACC & 71.01 & 71.22 & 67.90 & 70.11 \\
    \midrule
    \multirow{2}{*}{Flowers} & cACC & 80.21 & 76.57 & 83.59 & 77.86 \\
     & sACC & 56.42 & 69.78 & 42.86 & 61.49 \\
    \midrule
    \multirow{2}{*}{Pets} & cACC & 86.57 & 87.25 & 81.03 & 81.23 \\
     & sACC & 83.74 & 84.68 & 81.49 & 83.20 \\  
    \midrule
    \multirow{2}{*}{\textit{Average}} & cACC & 67.94 & 67.06 & 67.32 & 66.27 \\
     & sACC & 70.51 & 73.52 & 63.30 & 71.03 \\       
    \bottomrule
    \end{tabular}
    \end{adjustbox}
    \vspace{0.08in}
    \caption{Performance of public and private LMMs on five fine-grained benchmarks.
    Clustering accuracy (cACC) and semantic accuracy (sACC) are reported with (\checkmark) and without (\(\times\)) using a prompt augmented with indirect prior meta-data about the desired structure of the outputs. 
    It can be observed that an open-source LMM, when provided with our proposed advanced explicit reasoning techniques, can achieve similar or even better performance compared to the closed-source counterpart with proprietary built-in reasoning mechanisms.
    %
    %
    }    
    \label{tab:public_vs_private}
    \vspace{-0.1in}
\end{table}

Such a difference can be explained by Qwen2.5-VL's reliance on prompt engineering to exhibit reasoning, whereas Gemini2.5 comes with built-in thinking capabilities ``out-of-the-box". The proprietary model’s internal chain-of-thought mechanism is configurable via its API, but this advanced feature is locked behind a paywall. In practice, this means the private LMM can leverage sophisticated reasoning with minimal user prompting, while the public model requires carefully crafted prompts to reach similar levels of performance.
Notably, there are no open models with native internal reasoning yet, underscoring the trade-off between accessibility and built-in capability in current LMMs.

\subsubsection{Improving Label Semantics.}

Additionally, we analyse the performance of LMMs when given indirect prior meta-data about the desired structure of the outputs.
For example, the model can be provided with high-level meta-instructions suggesting it to prefer a more common flower's name or a car model name used for the North American market.
This setup could be useful for cases where obtaining the exact class names is more important than clustering functionality.
In Table \ref{tab:public_vs_private}, we report the results of such an evaluation on both public Qwen2.5-VL and private Gemini2.5-Flash large multi-modal models. 
From the result, it can be observed that with the dataset-specific instructions, our FiNDR shows significantly higher semantic accuracy (70.51 \% $\rightarrow$ 73.52 \% for Qwen2.5, and 63.30 \% $\rightarrow$ 71.03 \% for Gemini2.5) while the clustering metric decreases (67.94 \% $\rightarrow$ 67.06 \% for Qwen2.5, and 67.32 \% $\rightarrow$ 66.27 \% for Gemini2.5).
Remarkably, on the Flowers dataset, Qwen’s sACC rises by nearly 14 points, whereas its cACC drops by only 3.64 points.
Similarly, although less pronounced, these trends appear for Birds and Cars. 
In this way, enforcing stricter lexical conventions encourages the model to draw more high-level taxonomic distinctions, which may fragment visually coherent clusters.

This again challenges the assumption that human-curated vocabularies always define an actual upper bound.
While prompt engineering remains an effective lever for improving label semantics, it introduces a measurable trade-off in clustering precision.  
Future benchmarks could mitigate this tension by accepting synonym sets or taxonomic identifiers, enabling a more complete assessment of semantic fidelity.

\begin{table*}[!h]
\centering
\resizebox{\textwidth}{!}{
    \begin{tabular}{@{}lllllc@{}}
    \toprule
    \multirow{3}{*}{}           & \multicolumn{4}{|c|}{\textbf{One-time}}                                                                                      & \textbf{Multi-time}                             \\ \cmidrule(l){2-6} 
                                & \multicolumn{3}{|c|}{Discovery}                                 & \multicolumn{1}{c|}{\multirow{2}{*}{Refinement}} & \multirow{2}{*}{Inference}             \\ \cmidrule(lr){2-4}
                                & \multicolumn{1}{|l}{\textit{Meta}}         & \textit{Main}        & \multicolumn{1}{l|}{\textit{Service}}      & \multicolumn{1}{c|}{}                            &                                        \\ \midrule
    \multicolumn{1}{l|}{TFLOPs} & 32.62\(\times N_{img}\) & 5.13\(\times N_{img}\) & \multicolumn{1}{l|}{22.05\(\times N_{img}\)} & \multicolumn{1}{l|}{0.596 + 0.162\(\times N_{img}\)}          & \multicolumn{1}{l}{0.596 + 0.017\(\times N_{img}\)} \\
    \midrule
    \multicolumn{1}{l|}{time, ms}                    & 209\(\times N_{img}\)   & 32\(\times N_{img}\)   & \multicolumn{1}{l|}{141\(\times N_{img}\)}                        & \multicolumn{1}{l|}{3 + 1\(\times N_{img}\)}                                       & \multicolumn{1}{l}{3 + 0.1\(\times N_{img}\)}       \\ \bottomrule
    \end{tabular}
}
\caption{Approximate computational costs of our proposed FiNDR, including the one-time discovery and refinement steps and the multi-time inference step.
Average amount of FP16 computations required for a forward pass.
In the discovery step, the Qwen2.5-VL-72B-Instruct model with an average context length of 1,024 tokens is utilised.
For visual encoders, the input image is resized to 224$\times$224.
The processing time is calculated for an NVIDIA A100 GPU.
}
\label{tab:app_compute_cost}
\end{table*}

\begin{figure}[!h]
    \centering
    \includegraphics[width=1.0\columnwidth]{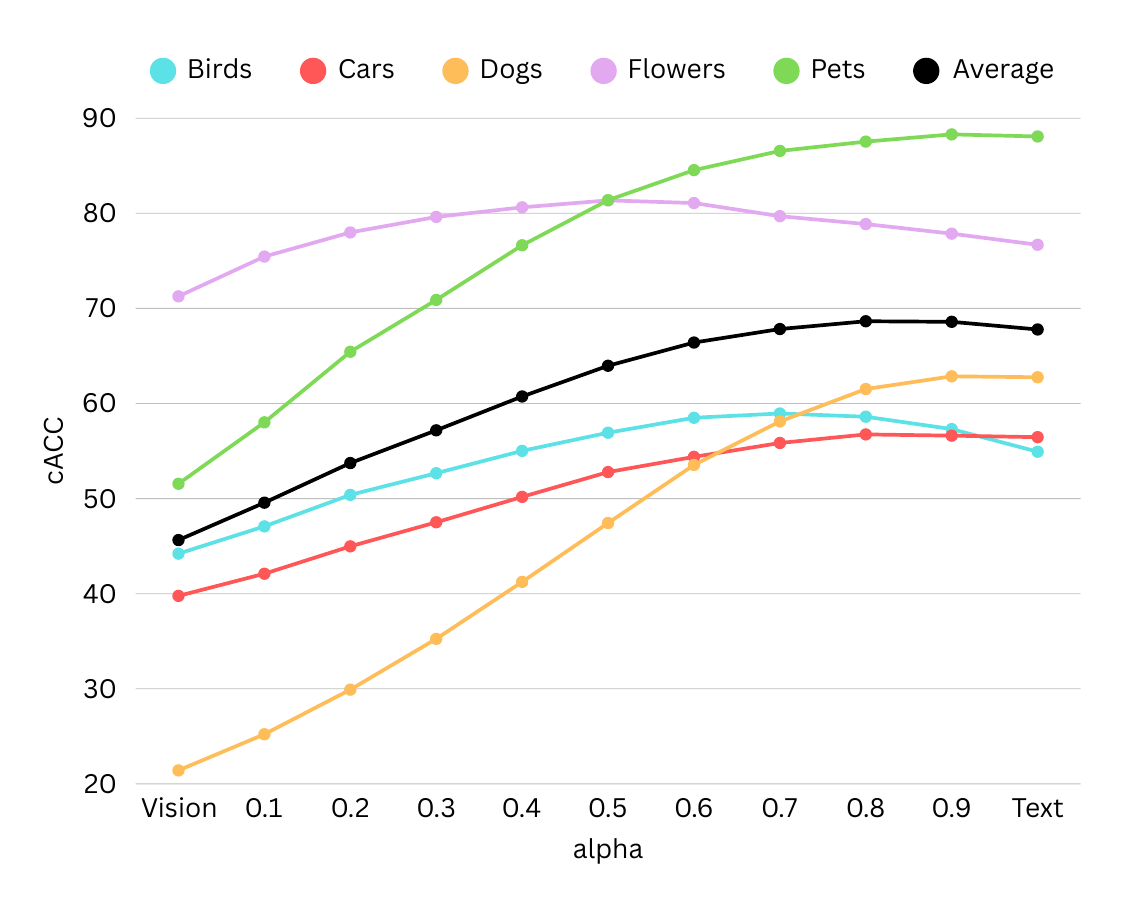}
    \vspace{-2.5em}
    \caption{Analysis of the $\alpha$ coefficient (Eq. 5) sensitivity based on clustering accuracy (cACC). The coefficient controls the contribution of visual and text components in the final class embedding.
    }
    \label{fig:app_alpha_ablation_cacc}
    \vspace{-0.5em}
\end{figure}

\begin{figure}[!h]
    \centering
    \includegraphics[width=1.0\columnwidth]{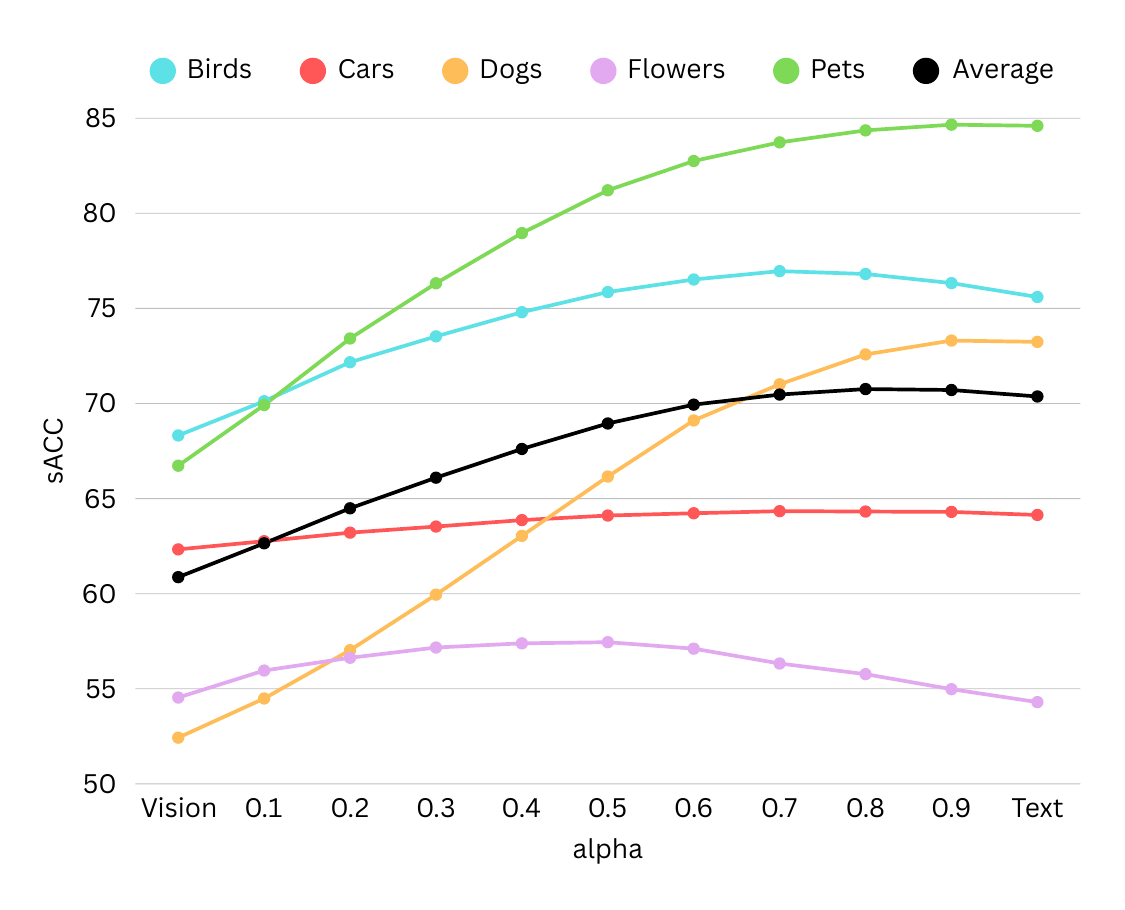}
    \vspace{-2.5em}
    \caption{Analysis of the $\alpha$ coefficient (Eq. 5) sensitivity based on semantic accuracy (sACC). The coefficient controls the contribution of visual and text components in the final class embedding.
    }
    \label{fig:app_alpha_ablation_sacc}
    \vspace{-0.5em}
\end{figure}

\subsubsection{Computational Costs Analysis.}
\label{app:compute_cost}


In Table \ref{tab:app_compute_cost}, we provide approximate computational costs of our proposed FiNDR, including the one-time discovery and refinement steps and the multi-time inference step.
In the discovery step, the average amount of FP16 computations required for a forward pass of the utilised Qwen2.5-VL-72B-Instruct model is $\approx$0.150 TFLOPs ($\approx$150.32 GFLOPs) per generated token (with average context length of 1,024 tokens).
For visual encoders, the input image is resized to 224$\times$224.
Additionally, approximate processing time is calculated for an NVIDIA A100 GPU.

It can be observed that the most noticeable amount of computations is only done one-time (discovery and refinement stages), while the multi-time inference compute cost remains highly efficient.
In this case, our FiNDR provides significantly higher accuracy while maintaining the same inference cost as prior SOTA.

\subsubsection{Analysis of $\alpha$ Hyper-parameter.}
\label{app:alpha_ablation}

In order to justify the choice of $\alpha$ coefficient in Equation~5, we conduct an analysis of the $\alpha$ hyper-parameter sensitivity based on
clustering accuracy (cACC) and semantic accuracy (sACC). We vary the value of $\alpha$ from 0.0 (vision-only mode) to 1.0 (text-only mode), thereby adjusting the contribution of the vision and text components to the final embedding. Each data point is obtained from one run.

As shown in Figures \ref{fig:app_alpha_ablation_cacc} and \ref{fig:app_alpha_ablation_sacc}, the textual component significantly improves the metrics in visually complex datasets.
As can be concluded, the right balance between the vision and text components enables the model to mitigate the potential domain shift and noisiness of guessed class names.

The value of $\alpha$ is expected to be adapted depending on the data domain, visual complexity, and class granularity.
In our experiments, we set $\alpha=0.7$ in order to maintain a balance between visual and textual data. This choice allows more space for error in the textual part, assuming that the visual component is always correct, while a discovered class name can potentially be wrong or noisy.

\subsubsection{Robustness Analysis.}
\label{app:domain_shift}



In order to assess the robustness of our FiNDR, we conduct multiple simulated scenarios where the predicted class names may be inaccurate.
Specifically, we consider 3 cases: a generic prediction, a misprediction, and random noise.
For this, we gradually replace a portion of the predicted class names with the specified failed prediction.

\begin{figure}[!h]
    \vspace{-0.5em}
    \centering
    \includegraphics[width=1.0\columnwidth]{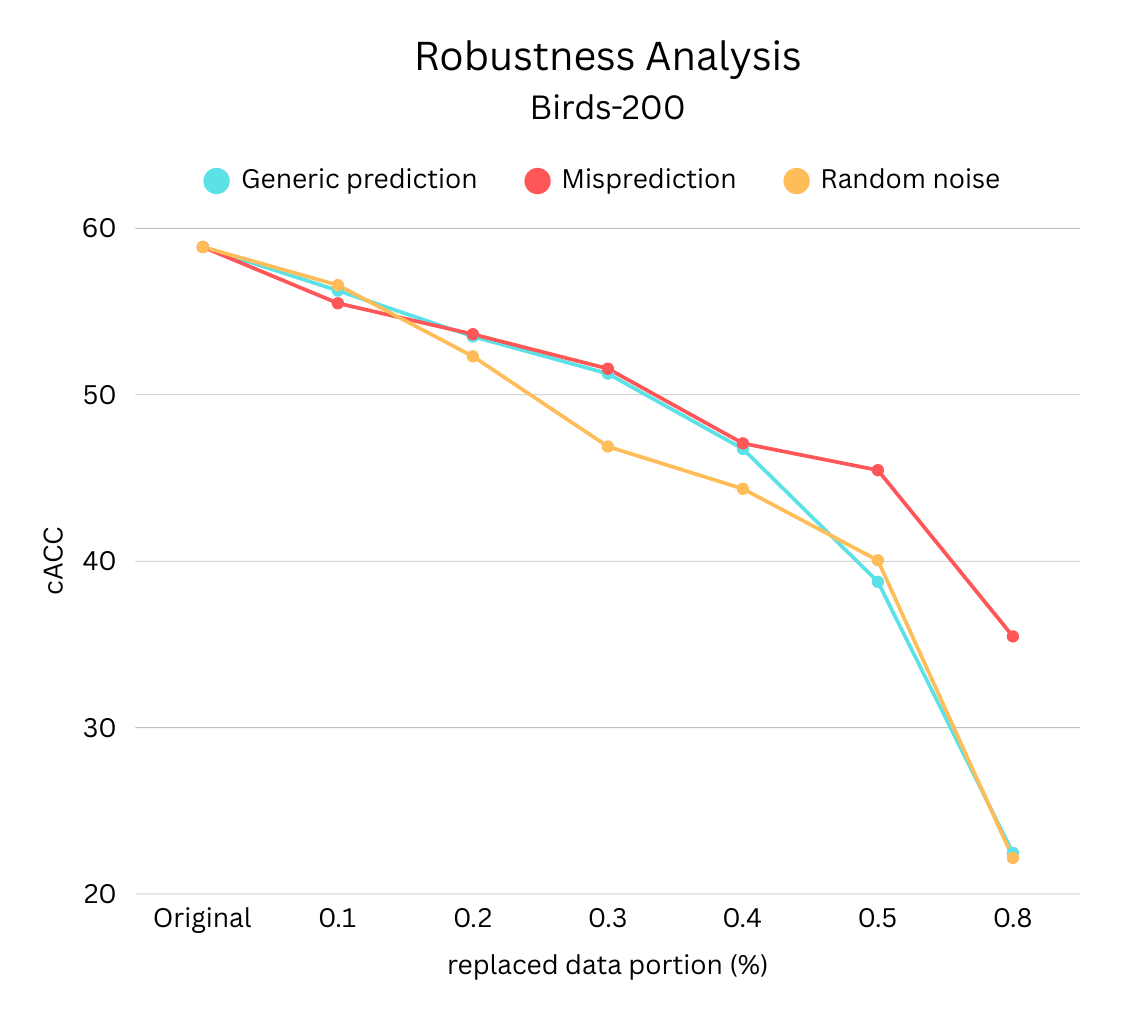}
    \caption{Analysis of our FiNDR's robustness under most common failure scenarios at the class name discovery stage: a generic prediction, a misprediction, and random noise.
    }
    \label{fig:app_robustness}
\end{figure}

As demonstrated in Figure \ref{fig:app_robustness}, our proposed pipeline remains robust even in the most challenging scenarios when 50\% of the predicted class names are inaccurate or corrupted.
This is largely attributed to the balanced choice of the $\alpha$ coefficient value in Equation 5 for vision-language mixing, which allows the model to leverage the complementary strengths of both modalities.




\end{document}